%% file: main.tex
\RequirePackage{fix-cm}
\documentclass[twocolumn]{svjour3}          
\smartqed  
\usepackage{graphicx}
\usepackage{amsmath}
\usepackage{xcolor}
\usepackage{amssymb}
\usepackage{multirow}
\usepackage{booktabs}
\usepackage{listings}
\usepackage{bm}
\usepackage{subfig}
\usepackage{hyperref}
\usepackage{appendix}
\usepackage{stfloats}
\usepackage{xurl}

\begin{document}

\title{View-Invariant, Occlusion-Robust Probabilistic Embedding for Human Pose
}


\author{
Ting~Liu$^{1,*}$ \and
Jennifer~J.~Sun$^{2,*}$
\and
Long~Zhao$^{1,3}$ \and 
Jiaping~Zhao$^1$ \and
Liangzhe~Yuan$^1$ \and
Yuxiao~Wang$^1$ \and
Liang-Chieh~Chen$^1$ \and
Florian~Schroff$^1$ \and 
Hartwig~Adam$^1$
}

\authorrunning{T.\ Liu, J.J.\ Sun et al.} 

\institute{
Ting~Liu \\
\email{liuti@google.com}\\
\\
Jennifer~J.~Sun\\
\email{jjsun@caltech.edu}\\
\\
Long~Zhao\\
\email{lz311@cs.rutgers.edu}\\
\\
Jiaping~Zhao\\
\email{jiapingz@google.com}\\
\\
Liangzhe~Yuan\\
\email{lzyuan@google.com}\\
\\
Yuxiao~Wang\\
\email{yuxiaow@google.com}\\
\\
Liang-Chieh~Chen\\
\email{lcchen@google.com}\\
\\
Florian~Schroff\\
\email{fschroff@google.com}\\
\\
Hartwig~Adam\\
\email{hadam@google.com}\\
\\
$^1$ Google Research, CA, USA.\\
$^2$ California Institute of Technology, CA, USA.\\
$^3$ Rutgers University, NJ, USA.\\
\\
$^*$ Equal contribution.
}

\date{Received: date / Accepted: date}

\maketitle

\input{0.abstract}

\input{1.introduction}

\input{2.related_work}

\input{3.pose_embedding}

\input{4.keypoint_dropout}

\input{5.experiments}

\input{6.conclusion}

\input{7.acknowledgements}


%
%

\bibliographystyle{spmpsci}      
\bibliography{references}   


\clearpage

\input{a1.appendix}

\end{document}

%% file: 0.abstract.tex
\begin{abstract}

Recognition of human poses and actions is crucial for autonomous systems to interact smoothly with people. However, cameras generally capture human poses in 2D as images and videos, which can have significant appearance variations across viewpoints that make the recognition tasks challenging. To address this, we explore recognizing similarity in 3D human body poses from 2D information, which has not been well-studied in existing works. Here, we propose an approach to learning a compact view-invariant embedding space from 2D body joint keypoints, without explicitly predicting 3D poses.
Input ambiguities of 2D poses from projection and occlusion are difficult to represent through a deterministic mapping, and therefore we adopt a probabilistic formulation for our embedding space.
Experimental results show that our embedding model achieves higher accuracy when retrieving similar poses across different camera views, in comparison with 3D pose estimation models.
We also show that by training a simple temporal embedding model, we achieve superior performance on pose sequence retrieval and largely reduce the embedding dimension from stacking frame-based embeddings for efficient large-scale retrieval.
Furthermore, in order to enable our embeddings to work with partially visible input, we further investigate different keypoint occlusion augmentation strategies during training. We demonstrate that these occlusion augmentations significantly improve retrieval performance on partial 2D input poses. Results on action recognition and video alignment demonstrate that using our embeddings without any additional training achieves competitive performance relative to other models specifically trained for each task.
\keywords{Human Pose Embedding \and Probabilistic Embedding \and View-Invariant Pose Retrieval \and Action Retrieval \and Occlusion Robustness}
\end{abstract}

%% file: 1.introduction.tex
\begin{figure}
  \centering
  \includegraphics[width=\columnwidth]{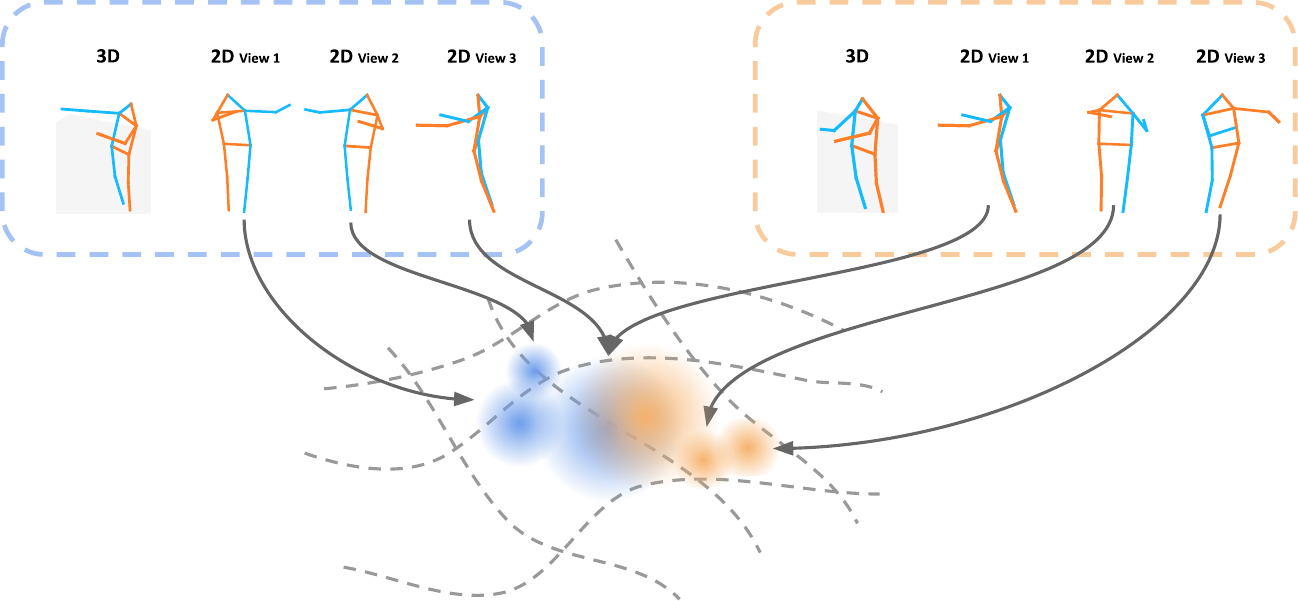}
  \caption{We embed 2D poses such that our embeddings are view-invariant (2D projections of similar 3D poses are embedded close together) and probabilistic (embeddings are distributions that cover different 3D poses projecting to the same input 2D pose).}
 \label{fig:introduction}
\end{figure}

\section{Introduction}

Automated perception of human poses and activities is an important step towards human-centric image and video understanding, which is crucial for applications such as autonomous vehicles~\cite{gu2019efficient}, and social robotics~\cite{garcia2019human}. In these applications, perception is often based on monocular cameras, which depict humans in two dimensions (2D). However, when we represent three-dimensional (3D) human poses in 2D, the same pose appears different across camera views due to changing relative depth of body parts and occlusions. These variations in 2D representations across viewpoints present a challenge for analyzing human behaviors from 2D data. To address this challenge, we explore using learning-based models to recognize similarity of 3D poses using 2D inputs.

We propose to learn view-invariant embeddings for 2D poses, which has not been well-explored in existing works. Typically, embedding models are trained from images using metric learning techniques~\cite{mori2015pose,ho2019pies,chu2019vehicle}. However, images with similar human poses can appear different due to a variety of factors, such as viewpoints, subjects, backgrounds, and clothing. As a result, it can be difficult to understand errors in the embedding space from a specific factor of variation. Furthermore, multi-view image datasets for human poses are difficult to capture in the wild with 3D groundtruth poses. Here, we embed 2D pose keypoints to view-invariant embeddings as illustrated in Fig.~\ref{fig:introduction}. Our method leverages existing 2D keypoint detectors, similar to 2D-to-3D lifting models~\cite{martinez2017simple,zhao2019semantic}. Using 2D keypoints as inputs enables our embedding model to focus on learning view-invariant pose features, and training with datasets captured in lab environments, while generalizing to in-the-wild data.

Another aspect of embedding models we aim to address is occlusion-robustness. For models to be applied in the wild, they need to be able to handle pose occlusions that are widely present in natural images. This ability to embed partial poses is also useful for searching for poses through parts. One simple solution is to train an individual embedding model for each possible occlusion pattern. However, this is infeasible in practice due to the large diversity of natural pose occlusions. Additionally, since these models are often trained using data captured in lab settings, the diversity of occlusion patterns for training is limited. In this paper, we propose to learn a unified occlusion-robust embedding model by synthesizing keypoint visibilities during training. We explore different synthetic keypoint dropout strategies to enable our model to be robust to common patterns of missing keypoints in-the-wild.

We additionally explore embedding temporal pose sequences, as temporal information is crucial for understanding human actions in videos.
One simple solution is to stack frame-level embeddings, but this results in high-dimensional embeddings that are prohibitive for some real-world applications (e.g., large-scale retrieval). On the other hand, learning to directly embed temporal sequences can potentially reduce the embedding dimensionality, and enables applications based on efficient sequence matching.

Finally, we study using probabilistic embeddings to address the input ambiguity of 2D poses. Many valid 3D poses can be projected to the same or very similar 2D pose~\cite{akhter2015pose}. This input uncertainty is difficult to represent using deterministic mappings to the embedding space (point embeddings)~\cite{oh2018modeling,kendall2017uncertainties}. Instead, we adopt probabilistic embeddings as multivariate Gaussian distributions and show that the learned variance from our method correlates with input 2D ambiguities.

One direct application of our embeddings is pose-based image retrieval. Our embeddings enable users to search images by fine-grained fully or partially visible pose, such as jumping with both hands up, running with one hand waving, and many other actions that are potentially difficult to predefine. The importance of this application is further highlighted by works such as~\cite{mori2015pose,jammalamadaka2012video}. Compared with using 3D keypoints with alignment for retrieval, our pose embeddings enable efficient similarity comparisons in Euclidean space. Our embeddings can also be applied to other tasks where recognizing pose similarity across views is important. Here, we demonstrate the performance using our embeddings on action recognition~\cite{zhang2013actemes,iqbal2017pose} and video alignment~\cite{dwibedi2019temporal} downstream tasks.

\paragraph{Contributions} We have developed a framework for mapping 2D poses to probabilistic embeddings where: (1) 2D pose embedding distances correspond to their similarities in absolute 3D pose space; (2) probabilistic embeddings capture uncertainty; (3) a single model embeds different pose visibility patterns for handling partial input. We evaluate our embeddings on cross-view pose retrieval, action recognition, and video alignment tasks.

This work features several extensions to our previous conference publication~\cite{sun2019view}, which focused on embedding fully visible 2D poses from a single frame.
First, we extend our single-frame embedding framework to handle sequential 2D inputs for cross-view sequence matching.
Second, we extend our framework to handle partially visible 2D input, which allows our embeddings to be more applicable to in-the-wild data, and enables users to define keypoint subsets for retrieval. We achieve this using a visibility mask and random keypoint dropouts during training, and we investigate the effect of different dropout strategies, including structured dropout that follows a prior distribution of realistic occlusions.
Further, we develop a benchmark for partial keypoint retrieval, for both natural occlusions and targeted pose retrieval, where users can specify retrieval of partial matches to an input 2D pose. Finally, we present a more extensive evaluation of our model and baselines, including comparing to more 3D pose estimation methods~\cite{zhao2019semantic,kocabas2019self} on more datasets~\cite{vonMarcard2018}.

Our code is released at {\small\url{https://github.com/google-research/google-research/tree/master/poem}}.

%% file: 2.related_work.tex
\section{Related Work}

We embed 2D human poses such that embedding space distance corresponds to 3D pose similarity. In this section, we review relevant literature on metric learning, human pose estimation, and view invariance and object retrieval.

\paragraph{Metric Learning} We are working to understand similarity in human poses across views. 
To capture measures of similarity between inputs, contrastive loss~\cite{bromley1994signature,hadsell2006dimensionality,oh2018modeling,oord2018representation,chen2020simple} or triplet loss (based on tuple ranking)~\cite{wang2014learning,schroff2015facenet,wohlhart2015learning,hermans2017defense} are commonly used. These losses are used to push together examples that are similar in the embedding space and pull apart examples that are dissimilar. In visual representation and metric learning~\cite{oh2018modeling,oord2018representation,chen2020simple,schroff2015facenet}, this similarity generally corresponds to categorical image class labels. Our work is different in that our similarity measure is based on continuous 3D pose distance and we embed 2D pose keypoints, which allows us to explore distinct approaches for pose representations.

In our work, we learn a mapping from Euclidean distance in the embedding space to a probabilistic pose similarity score. This probabilistic similarity captures closeness in 3D pose space from 2D poses. Our work is inspired by the mapping used in soft contrastive loss~\cite{oh2018modeling} for learning from an occluded N-digit MNIST dataset. 

During training of contrastive loss or triplet loss, the number of possible training tuples increases exponentially with respect to the number of samples in the tuple, and not all combinations are equally informative.
To find informative training tuples, various mining strategies are proposed~\cite{schroff2015facenet,wu2017sampling,oh2016deep,hermans2017defense}. In particular, semi-hard triplet mining has been widely used~\cite{schroff2015facenet,wu2017sampling,parkhi2015deep}. We measure the hardness of a negative sample based on its embedding distance to the anchor. Commonly, this distance is the Euclidean distance~\cite{wang2014learning,wohlhart2015learning,schroff2015facenet,hermans2017defense}, but any differentiable distance function could be applied~\cite{hermans2017defense}. In particular, \cite{huang2016local,iscen2018mining} show that alternative distance metrics also work for image and object retrieval. In our work, the distance metric is based on a probabilistic 3D pose similarity score from embedded 2D poses.

Most of the papers discussed above involve deterministically mapping inputs to point embeddings. We map inputs to probabilistic embeddings, similar to works such as~\cite{vilnis2014word,bojchevski2017deep,oh2018modeling}. Probabilistic embeddings have been used to model specificity of word embeddings~\cite{vilnis2014word}, uncertainty in graph representations~\cite{bojchevski2017deep}, and input uncertainty due to occlusion~\cite{oh2018modeling}. We will apply probabilistic embeddings to address inherent ambiguities in 2D pose due to 3D-to-2D projection.

\paragraph{Human Pose Estimation} 3D human pose estimation from monocular 2D input, such as images or 2D poses, is a widely explored area~\cite{martinez2017simple,chen20173d,pavllo20193d,rayat2018exploiting,zhou2017towards,sun2018integral,rhodin2018unsupervised,tekin2017learning,chen2019unsupervised,rhodin2018learning}. Instead of mapping 2D poses to 3D poses, our work maps 2D poses to a view-invariant embedding space. This embedding space enables detected 2D poses to be matched across views and is directly applicable to pose retrieval, action recognition, and video alignment. 

Many approaches estimate 3D poses in the camera coordinate system~\cite{martinez2017simple,chen20173d,pavllo20193d,rayat2018exploiting,zhou2017towards,sun2018integral,rhodin2018unsupervised,tekin2017learning,chen2019unsupervised,li2020cascaded,zeng2020srnet}. Since the pose description changes based on camera viewpoint, 3D poses in the camera coordinate frame is not view-invariant, and cross-view retrieval requires rigid alignment for every pose pair.

2D-to-3D lifting pose estimators~\cite{martinez2017simple,chen20173d,pavllo20193d,rayat2018exploiting,zhao2019semantic} use detected 2D poses as inputs, similar to our approach. Some of these works~\cite{li2020cascaded,zeng2020srnet} also use data augmentation to improve generalization to new poses for 3D lifting.
Lifting models are trained to regress to 3D pose keypoints, while our model is trained using metric learning and outputs an embedding distribution. In addition, other works use multi-view datasets to predict 3D poses in the global coordinate frame~\cite{qiu2019cross,kocabas2019self,iskakov2019learnable,rhodin2018learning,tome2018rethinking}. Our work differs from these methods with our goal of learning view-invariant embeddings, approach of metric learning, and downstream tasks.

Although our approach focuses on learning pose embeddings, it is worth discussing 3D pose estimation models that also take occlusion into account during training, notably~\cite{sarandi2018synthetic,cheng2019occlusion,cheng20203d}. \cite{sarandi2018synthetic} applies occlusion augmentation to images by randomly applying patches to pose images. \cite{cheng2019occlusion} uses 3D poses to estimate occlusion patterns due to self-occlusion, by approximating a human body with cylinders. Finally, \cite{cheng20203d} applies random and area-based occlusion patterns to detected keypoint heatmaps.  
While we also experiment with random occlusion dropout, we additionally compute a prior distribution of body occlusions from on a large amount of random photos from the Internet. This enables us to use realistic distributions to augment our model training. As a result, our model performs better on these occlusion patterns.

\paragraph{View Invariance} When we capture a 3D scene in 2D as images or videos, changing the viewpoint often does not change other properties of the scene. To the best of our knowledge, we are the first to explore mapping 2D human poses to a view-invariant embedding space via metric learning. View invariance from 2D information can enable a variety of vision applications, such as motion analysis~\cite{ji2008visual,ji2009advances,ronchi2016rotation}, video alignment~\cite{dwibedi2019temporal}, tracking~\cite{ong2006viewpoint}, vehicle and human re-identification~\cite{chu2019vehicle,zheng2019pose}, object classification and retrieval \cite{lecun2004learning,hu2010learning,ho2019pies}, and action recognition~\cite{rao2001view,liu2018viewpoint,xia2012view,li2018unsupervised}. These downstream tasks can potentially benefit from our view-invariant pose embeddings. Here, we investigate how view invariance can be applied to cross-view retrieval, action recognition, and video alignment.

\begin{figure*}
  \centering
  \includegraphics[width=0.9\textwidth]{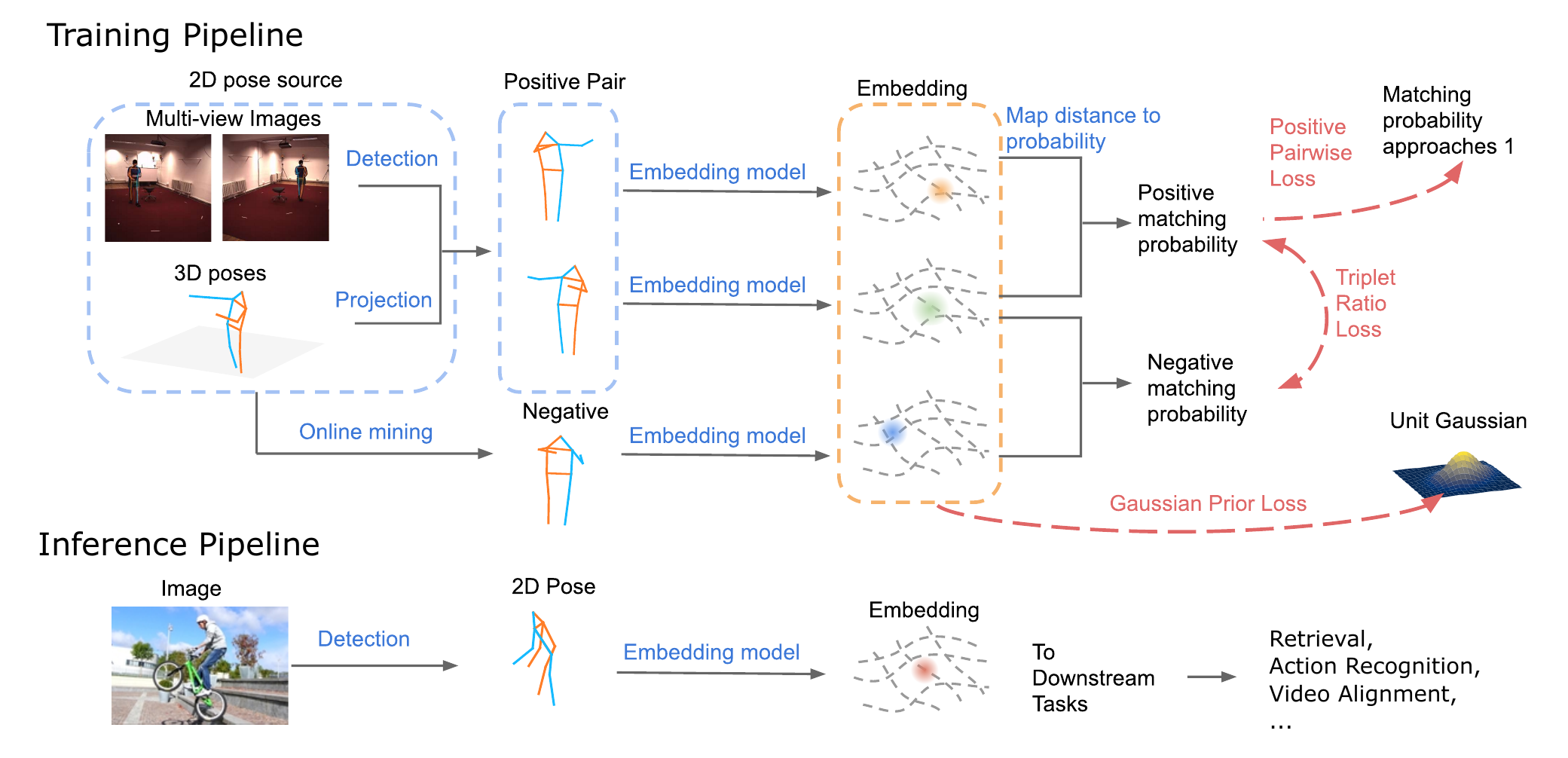}
  \caption{Overview of Pr-VIPE model training and inference. Our model takes keypoint input from a single 2D pose (detected from images and/or projected from 3D poses) and predicts embedding distributions. Three losses are applied during training.}
 \label{fig:methods}
\end{figure*}

\paragraph{Object Retrieval} With growing amounts of recorded data, retrieval has received increasing amounts of attention in research communities ~\cite{lecun2004learning,mori2015pose,hu2010learning,ho2019pies,he2018triplet}. The ability to retrieve similar images and videos according to different similarity metrics is useful for a variety of vision applications. Here, we would like to retrieve images containing similar 3D poses using 2D information. We show that this ability also enables us to achieve cross-view action recognition and video alignment. 

Retrieval of 2D poses in the same view has been studied in~\cite{mori2015pose}. This method embeds images with similar 2D poses in the same view close together but is not view-invariant. Our method focuses on learning view invariance, and we also differ from~\cite{mori2015pose} in method, using probabilistic 2D pose embeddings.

Compared to classification-based retrieval tasks, such as~\cite{ho2019pies}, our domain differs in a few ways. Our targets are continuous 3D poses, whereas in object recognition and retrieval tasks, each embedding is associated with a discrete class label. Furthermore, we embed 2D poses instead of images. Our approach allows us to investigate the impact of input 2D uncertainty with probabilistic embeddings and explore different methods to measure cross-view pose retrieval confidence. We hope that our work provides a novel perspective on view invariance for human poses.

%% file: 3.pose_embedding.tex
\section{View-Invariant Probabilistic Pose Embedding}\label{sec:pr_vipe}

Our goal is to embed 2D poses such that distances in the embedding space correspond to similarities of their corresponding absolute 3D poses in the Euclidean space. We achieve this view-invariance property via training with the triplet ratio loss (Section~\ref{sec:triplet_ratio_loss}), which pulls together 2D poses corresponding to similar 3D poses and pushes apart 2D poses corresponding to dissimilar 3D poses. The positive pairwise loss (Section~\ref{sec:positive_pairwise_loss}) is applied to increase the matching probability of similar poses. Finally, the Gaussian prior loss (Section~\ref{sec:probabilistic_embeddings}) helps regularize embedding magnitude and variance. The training and inference framework of our model is illustrated in Fig.~\ref{fig:methods}.

\subsection{Matching Definition}\label{sec:matching_definition}

We define a measure of similarity so that we can pull together 2D poses corresponding to similar 3D poses. The 3D pose space is continuous, and two 3D poses can be trivially different without being identical. To account for this, we define two 3D poses to be matching if they are visually similar regardless of viewpoint.
Given two sets of 3D pose keypoints $(\bm{y}_i,\bm{y}_j)$, we define a matching indicator function
\begin{equation} \label{eq:12}
 \vspace{-0.1cm}
m_{ij} = \begin{cases}
    1, & \text{if } \text{NP-MPJPE}(\bm{y}_i, \bm{y}_j) \leqslant \kappa\\
    0,              & \text{otherwise,}
\end{cases}
\end{equation}
where $\kappa$ controls visual similarity between matching poses.
Here, we use the mean per joint position error (MPJPE)~\cite{ionescu2013human3} between the two sets of 3D pose keypoints as a proxy to quantify their visual similarity. Before computing MPJPE, we normalize the 3D poses, as described in Section~\ref{sec:pose_normalization}, and apply Procrustes alignment between them. We do this in order for our model to be view-invariant and to disregard rotation, translation, or scale differences between 3D poses. We refer to this normalized, Procrustes-aligned MPJPE as \textbf{NP-MPJPE}.

\begin{figure}
\centering
\includegraphics[width=\columnwidth]{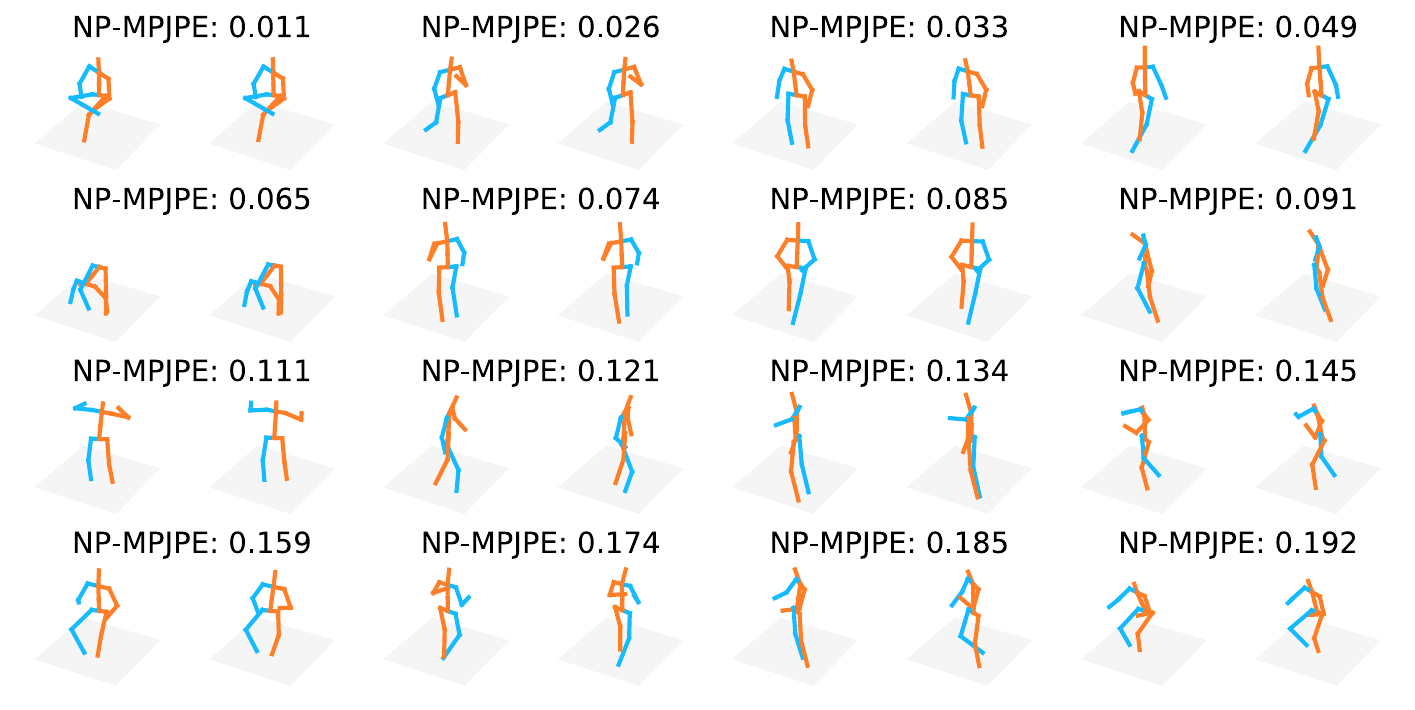}
\caption{3D pose pairs with different NP-MPJPEs, where the NP-MPJPEs increase with each row. The poses are randomly sampled from the hold-out set of H3.6M. Row 1 shows pairs with $0$ to $0.05$ NP-MPJPEs, row 2 shows pairs with $0.05$ to $0.1$ NP-MPJPEs, row 3 shows pairs with $0.1$ to $0.15$ NP-MPJPEs, and row 4 shows pairs with $0.15$ to $0.2$ NP-MPJPEs.}
\label{fig:similarity}
\end{figure}

Fig.~\ref{fig:similarity} demonstrates sampled 3D pose pairs with different ranges of corresponding NP-MPJPE between them. This plot shows the effect of choosing different $\kappa$. Unless stated otherwise, our models use $\kappa=0.1$, which corresponds to using the first two rows as matching pairs and the rest of the rows as non-matching. We note that pairs in rows 3 and 4 shows significant visual differences as compared with the first two rows. In general, $\kappa$ can be set by the user based on the perception of visual similarity.

\subsection{Triplet Ratio Loss}\label{sec:triplet_ratio_loss}
The triplet ratio loss aims to embed 2D poses based on the matching indicator function (\ref{eq:12}). Let $n$ be the dimension of the input 2D pose keypoints $\bm{x}$, and $d$ be the dimension of the output embedding. We would like to learn a mapping $f: \mathbb{R}^n \rightarrow \mathbb{R}^d$, such that:
\begin{equation}
D(\bm{z}_i, \bm{z}_j) < D(\bm{z}_i, \bm{z}_{j^{\prime}}),\forall m_{ij}>m_{ij^{\prime}},
\end{equation}
where $\bm{z} = f(\bm{x})$, 
and $D(\bm{z}_i, \bm{z}_j)$ is an embedding space distance measure.

For a pair of input 2D poses $(\bm{x}_i,\bm{x}_j)$, we define $p(m|\bm{x}_i,\bm{x}_j)$ to be the probability that their corresponding 3D poses $(\bm{y}_i,\bm{y}_j)$ match, that is, they are visually similar. While it is difficult to define this probability directly, we propose to assign its values by estimating $p(m|\bm{z}_i,\bm{z}_j)$ via metric learning. We know that if two 3D poses are identical, then $p(m|\bm{x}_i,\bm{x}_j)=1$, and if two 3D poses are sufficiently different, $p(m|\bm{x}_i,\bm{x}_j)$ should be small. 
For any given input triplet $(\bm{x}_i,\bm{x}_{i^+},\bm{x}_{i^-})$ with $m_{i,i^+}>m_{i,i^-}$, we want

\begin{equation}
\frac{p(m|\bm{z}_i, \bm{z}_{i^+})}{p(m|\bm{z}_i,\bm{z}_{i^-})}\geqslant\beta,
\end{equation}
where $\beta>1$ represents the ratio of the matching probability of a similar 3D pose pair to that of a dissimilar pair. That is, the matching probability of a positive pair should be larger than the matching probability of a negative pair.

Applying negative logarithm to both sides, we have
\begin{equation}
(-\log p(m|\bm{z}_i, \bm{z}_{i^+})) - (-\log p(m|\bm{z}_i, \bm{z}_{i^-}))\leqslant-\log\beta.
\end{equation}
We note that the training can optimize this using the triplet loss framework~\cite{schroff2015facenet}. Given batch size $N$, we define the triplet ratio loss $\mathcal{L}_\text{ratio}$ as
\begin{equation}\label{eq:ratio_loss}
\mathcal{L}_\text{ratio}=\sum_{i=1}^N\max(0,D_m(\bm{z}_i,\bm{z}_{i^+})-D_m(\bm{z}_i,\bm{z}_{i^-})+\alpha),
\end{equation}
with distance kernel $D_m(\bm{z}_i,\bm{z}_j)=-\log p(m|\bm{z}_i,\bm{z}_j)$ and margin $\alpha=\log\beta$. To form a triplet $(\bm{x}_i,\bm{x}_{i^+},\bm{x}_{i^-})$, we set the anchor $\bm{x}_i$ and positive $\bm{x}_{i^+}$ to be projected from the same 3D pose and perform online semi-hard negative mining~\cite{schroff2015facenet} to find $\bm{x}_{i^-}$.

We can compute the matching probability using our embeddings. To compute $p(m|\bm{z}_i, \bm{z}_j)$, we use the formulation proposed by~\cite{oh2018modeling}:
\begin{equation} \label{eq:6}
p(m|\bm{z}_i, \bm{z}_j) = \sigma(-a ||\bm{z}_i - \bm{z}_j||_2 + b),
\end{equation}
where $\sigma$ is a sigmoid function, and the trainable scalar parameters $a>0$ and $b\in\mathbb{R}$ calibrate embedding distance to probabilistic similarity. 

\subsection{Positive Pairwise Loss}\label{sec:positive_pairwise_loss}
Since the positive pairs in our triplets have identical 3D poses, we would like them to have high matching probabilities. We encourage this property by adding the positive pairwise loss

\begin{equation} \label{eq:7}
\mathcal{L}_\text{positive} = \sum_{i=1}^N - \log p(m|\bm{z}_i, \bm{z}_{i^+}).
\end{equation}

The combination of $\mathcal{L}_\text{ratio}$ and $\mathcal{L}_\text{positive}$ can be applied to training point embedding models, which we refer to as VIPE.

\subsection{Probabilistic Embeddings}\label{sec:probabilistic_embeddings}
In this section, we discuss the extension of VIPE to the probabilistic formulation Pr-VIPE. The inputs to our model, 2D pose keypoints, are inherently ambiguous, and there are many valid 3D poses that can be projected to a similar 2D pose~\cite{akhter2015pose}. This input uncertainty can be difficult to model using point embeddings~\cite{kendall2017uncertainties,oh2018modeling}. We investigate representing this uncertainty using distributions in the embedding space by mapping 2D poses to probabilistic embeddings: $\bm{x} \rightarrow p(\bm{z}|\bm{x})$. Similar to~\cite{oh2018modeling}, we extend the input matching probability~(\ref{eq:6}) to using probabilistic embeddings as $p(m|\bm{x}_i,\bm{x}_j)=\int p(m|\bm{z}_i,\bm{z}_j)p(\bm{z}_i|\bm{x}_i)p(\bm{z}_j|\bm{x}_j)\textrm{d}\bm{z}_i\textrm{d}\bm{z}_j$, which can be approximated using Monte-Carlo sampling with $K$ samples drawn from each distribution as
\begin{equation} \label{eq:9}
p(m|\bm{x}_i, \bm{x}_j) \approx \frac{1}{K^2} \sum_{k_1 = 1}^K \sum_{k_2 = 1}^K p(m|\bm{z}_i^{(k_1)}, \bm{z}_j^{(k_2)}).
\end{equation}

We model $p(\bm{z}|\bm{x})$ with a $d$-dimensional Gaussian distribution with a diagonal covariance matrix. The model estimates mean $\mu(\bm{x})\in\mathbb{R}^d$ and covariance $\Sigma(\bm{x})\in\mathbb{R}^d$ with a shared base network and separate output layers. During sampling, we use the reparameterization trick~\cite{kingma2013auto}, by first sampling from a unit Gaussian $\epsilon^{(k)}\sim\mathcal{N}(\bm{0},\bm{I})$ and then computing $\bm{z}^{(k)}=\mu(\bm{x})+\epsilon^{(k)}\cdot\textrm{diag}(\Sigma(\bm{x})^{1/2})$ during training for easy backpropagation.

We place a unit Gaussian prior on our embeddings with KL divergence by adding the Gaussian prior loss
\begin{equation} \label{eq:10}
\mathcal{L}_\text{prior} =\sum_{i=1}^N D_\text{KL}(\mathcal{N}(\mu(\bm{x}_i), \Sigma(\bm{x}_i))\:\|\: \mathcal{N}(\bm{0},\bm{I})).
\end{equation}
This loss prevents variance from collapsing to zero and regularizes embedding mean magnitudes

\subsection{Training}~\label{sec:training_objective}
Our full training objective for Pr-VIPE is:
\begin{equation} \label{eq:11}
\mathcal{L} = w_\text{ratio} \mathcal{L}_\text{ratio} + w_\text{positive} \mathcal{L}_\text{positive} + w_\text{prior} \mathcal{L}_\text{prior}.
\end{equation}
We optimize training with respect to $a$ and $b$ in~(\ref{eq:6}) and the embedding function. We set the loss weights such that all the terms have similar magnitude with $w_\text{ratio} = 1$.

We adopt the backbone model architecture in~\cite{martinez2017simple} with two residual fully-connected (FC) blocks, though we note that other architectures may also work. More implementation details can be found in Section~\ref{sec:training_details}.

\subsection{Pose Normalization}\label{sec:pose_normalization}

We normalize our 2D and 3D poses such that camera parameters are not needed during training and inference. 
For 3D poses, our normalization procedure is similar to that in~\cite{chen2019unsupervised}. We translate a 3D pose so that the pelvis located at the origin. We then scale the pelvis to spine to neck distance to a unit scale. 
For 2D poses, we translate the keypoints so that the center between left and right hip is at the origin. Then we normalize the pose such that the maximum distance between shoulder and hip joints is $0.5$. This maximum distance is computed between all pairwise distances among left shoulder, right shoulder, left hip, and right hip.

\subsection{Camera Augmentation}\label{sec:keypoint_augmentation}

During training, input triplets consist of detected and/or projected 2D keypoints as shown in Fig.~\ref{fig:methods}. When we train with detected 2D keypoints only, we are limited to the camera views in training images. In order to adapt our model to more views, we perform camera augmentation by generating triplets using 2D pose keypoints projected from 3D poses at random views. These triplets are then mixed with the triplets from 2D detection for training.

To form triplets using multi-view image pairs, we use detected 2D keypoints from different views as anchor-positive pairs.
To use projected 2D keypoints, we perform two random rotations to a normalized input 3D pose to generate two 2D poses from different views for anchor-positive pairs. Camera augmentation is then performed by using a mixture of detected and projected 2D keypoints.

\subsection{Temporal Pose Embedding}\label{sec:temporal_embeddings}

For understanding actions, sequences of human poses are usually required as they provide important temporal information. In this section, we further extend the Pr-VIPE framework to temporal domain, namely Temporal Pr-VIPE, to explicitly handle sequential inputs. Instead of embedding a single 2D pose, we embed a 2D pose sequence with the view-invariance and probabilistic properties of Pr-VIPE.

The input to our Temporal Pr-VIPE is a sequence of 2D poses from $T$ temporally-ordered frames. Atrous sampling is used with a rate based on the video frame rate. We then apply the full Pr-VIPE objective (Section~\ref{sec:training_objective}) to train an embedding model for 2D pose sequences. To compare whether a pair of pose sequences are similar, we compute the NP-MPJPEs of the 3D poses between each of their corresponding frame pairs and threshold on the maximum pairwise NP-MPJPE.
We also apply camera augmentation, similar to the Pr-VIPE training, by applying a random camera view to a subset of sequences within each batch during training.

\begin{figure}
\centering
\includegraphics[width=\columnwidth]{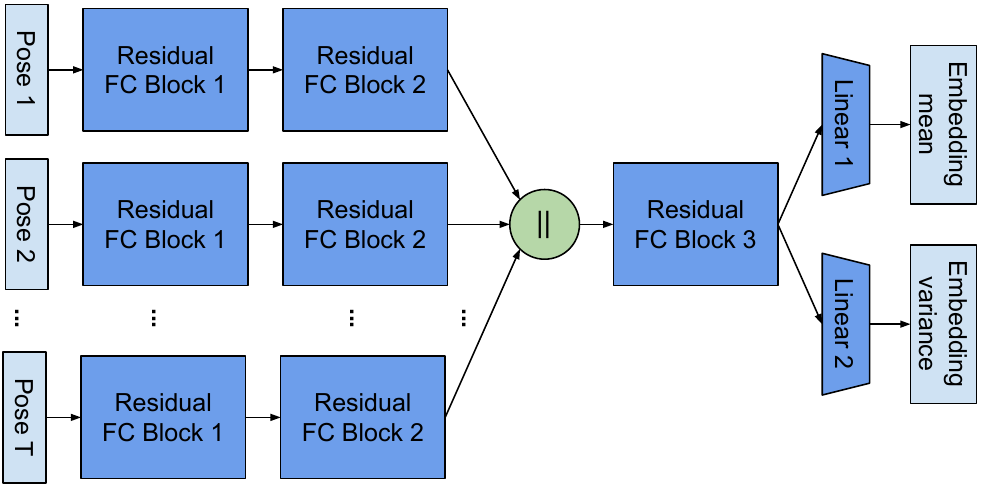}
\caption{Temporal Pr-VIPE model architecture. The green circle represents vector concatenation.}
\label{fig:temporal_pr_vipe_arch}
\end{figure}

We adopt a mid-fusion style network model architecture. Each 2D pose from the input sequence is fed into a network with two residual FC blocks~\cite{martinez2017simple}, and the output features are concatenated and fed to a third residual FC block followed by linear heads for final predictions, as shown in Fig.~\ref{fig:temporal_pr_vipe_arch}. More details on the implementation can be found in Section~\ref{sec:training_details}.

\subsection{Inference} At inference time, the Pr-VIPE model takes a single 2D pose (either from detection or projection) and outputs the mean and the variance of the embedding Gaussian distribution. The Temporal Pr-VIPE model takes in $T$ context frames, and outputs the mean and the variance of the embedding Gaussian distribution. We do not need explicit camera parameters for training or inference.

%% file: 4.keypoint_dropout.tex
\section{Occlusion-Robust Pose Embedding}\label{sec:occlusion_robust_embedding}

In this section, we extend the Pr-VIPE framework to handle inputs with different keypoint visibilities by embedding them into a single space using a single model.
We first present our keypoint visibility representation for modeling in Section~\ref{sec:keypoint_vis_representation}. We then study two keypoint occlusion augmentation strategies in Section~\ref{sec:keypoint_availability_aug}: independent keypoint dropout and structured keypoint dropout. Finally, we describe training with keypoint occlusions in Section~\ref{sec:keypoint_vis_training}.

\subsection{Keypoint Visibility Representation}\label{sec:keypoint_vis_representation}
We use a binary mask vector $\bm{v}$ to represent the visibility of each keypoint for an input 2D pose $\bm{x}$, where each entry of $\bm{v}$ is $1$ if its corresponding keypoint is visible and $0$ otherwise. This keypoint visibility indicator can represent whether a keypoint is invisible due to occlusion or being excluded. For example, keypoint masks can be used for retrieval of partial pose matches. When a keypoint has mask $0$, its normalized coordinates are also set to $0$. We concatenate the keypoint mask with normalized 2D keypoints as the model input.

In practice, we always consider the torso keypoints (two shoulders and two hips) as visible, based on the fact that torso has a near-rigid shape represented by these four keypoints, for which our 2D keypoint detector mostly gives reliable location estimates.

\subsection{Keypoint Occlusion Augmentation}\label{sec:keypoint_availability_aug}
In order for our model to be robust to partially visible input, we generate various keypoint occlusion patterns during training. It is ideal to train our model with diverse realistic keypoint occlusion patterns, which, however, our dataset does not provide. We address this by synthesizing occlusion patterns on fully visible poses for training. Here, we explore two methods for creating synthetic keypoint visibility masks:

\paragraph{Independent Keypoint Dropout} One simple way to synthesize keypoint occlusion patterns is to randomly mask keypoints during training. We apply an i.i.d.\ sampling with fixed chance $q$ to determine whether each keypoint of an input 2D pose is to be marked as invisible at each training iteration.

\paragraph{Structured Keypoint Dropout} In realistic photos, the visibilities of pose keypoints are usually not i.i.d. A more sophisticated way to synthesize ``realistic'' keypoint occlusion patterns is to consider the joint distribution of keypoint visibilities. We estimate 2D keypoints of about $300$ million pose samples detected from random in-the-wild photos. We binarize the keypoint detection confidence via thresholding and use it as a proxy to keypoint visibility. We then approximate the joint distribution of keypoint visibility patterns in the wild with its counting frequency.

The number of patterns in the joint distribution is exponential to the number of keypoints. We decompose the full body pose graph (See keypoint definitions in Appendix~\ref{sec:appendix_keypoint_definition}) into $6$ cliques:
\begin{itemize}
    \item Head: nose tip.
    \item Torso: left shoulder, right shoulder, left hip, right hip.
    \item Left arm/torso: left wrist, left elbow, left shoulder, left hip.
    \item Right arm/torso: right wrist, right elbow, right shoulder, right hip.
    \item Upper legs: left hip, right hip, left knee, right knee.
    \item Lower legs: left knee, right knee, left ankle, right ankle.
\end{itemize}
We sample the visibility mask values for each clique following this order. When sampling for a clique, if some of its keypoints have been assigned a visibility mask value, then we marginalize them from the joint distribution, and sample from the rest options. All of the cliques have at most four keypoints, and thus marginalizing and sampling from their joint distributions can be easily handled.

In practice during training, we first apply a threshold to keypoint detection confidence and mask out keypoints with low confidence. Then we further sample the keypoint mask values based on either strategies. 

\subsection{Training with Occlusions}\label{sec:keypoint_vis_training}
Following the Pr-VIPE triplet training framework, we generate keypoint visibility mask using the strategies mentioned above for all anchor poses and use fully visibility masks for their positive matches. Semi-hard negative matches are mined online from the batch and assigned full visibility masks. The negative labels are determined according to the 3D pose matching definition (Section~\ref{sec:matching_definition}) based on visible keypoints.

%% file: 5.experiments.tex
\section{Experiments}

In this section, we describe our experiment procedures and present our model evaluations on three tasks: cross-view retrieval, action recognition, and video alignment.
First, we show our model performance on pose retrieval across different camera views for fully-visible poses (Section~\ref{sec:retrieval_subsection}), pose sequences (Section~\ref{sec:sequence_retrieval}), and partially-visible poses (Section~\ref{sec:partial_retrieval}). Second, we show that our embeddings can be directly applied to action recognition (Section~\ref{sec:action_recognition}), without any additional training. Finally, we present the results of using our embeddings for video alignment (Section~\ref{sec:sequence_alignment}).

\subsection{Datasets}\label{sec:datasets}

For all the experiments in this paper, we only train on a subset of the Human3.6M~\cite{ionescu2013human3} dataset. For pose retrieval experiments, we validate on the Human3.6M hold-out set and test on a different dataset MPI-INF-3DHP~\cite{mehta2017monocular}, which is unseen during training and free from parameter tuning. To evaluate models for handling partially visible poses, we create a number of datasets based on the Human3.6M hold-out set with synthetically occluded keypoints. Additionally, we use the 3D Poses in the Wild dataset~\cite{vonMarcard2018}, which includes realistic keypoint occlusions.
We present qualitative results on MPII Human Pose~\cite{andriluka20142d}, for which 3D groundtruth is not available. For action recognition and video alignment experiments, we apply our model on Penn Action~\cite{zhang2013actemes}, a video action dataset unseen during training.

\paragraph{Human3.6M (H3.6M)}\cite{ionescu2013human3} H3.6M is a large human pose dataset recorded from $4$ chest-level cameras at $50$ frames per second (FPS) with 3D pose groundtruth. We follow the standard protocol~\cite{martinez2017simple}: train on Subject 1, 5, 6, 7, and 8, and hold out Subject 9 and 11 for validation. For evaluation, we remove near-duplicate 3D poses within $0.02$ NP-MPJPE, resulting in a total of $10910$ evaluation frames per camera. This process is camera-consistent, meaning if a frame is selected under one camera, it is selected under all cameras, so that the perfect retrieval result is possible. 
For temporal embeddings, we use the same dataset split and evaluation frames. In particular, we use $0.5$-second clips centered at the evaluation frames for the single-pose retrieval experiments.

Note that H3.6M is the only dataset we use for training our models in this paper. The following datasets are used {\bf only for testing and not for any training}:

\paragraph{MPI-INF-3DHP (3DHP)}\cite{mehta2017monocular} 3DHP is a more recent human pose dataset that contains $14$ diverse camera views and scenarios, covering more pose variations than H3.6M. We use $11$ cameras from this dataset and exclude the $3$ cameras with overhead views. 
Similar to H3.6M, we remove near-duplicate 3D poses, resulting in $6824$ frames per camera. We use all 8 subjects from the train split of 3DHP. 
For temporal embeddings, we use $0.5$-second clips centered at these $6824$ frames for the single-pose retrieval experiments. We note that videos from 3DHP are either $25$ or $50$ FPS.

\paragraph{3D Poses in the Wild (3DPW)}\cite{vonMarcard2018} 3DPW contains $60$ videos of human poses captured in the wild with realistic occlusions. We use this dataset to evaluate our occlusion-robust models and related baselines in terms of pose retrieval. We split the test set originally proposed in~\cite{vonMarcard2018} by randomly selecting $45\%$ of the video sequences for query samples, and the rest for index samples. We evenly downsample the video frames by $2$, and only include a query sample if (1) its four 2D torso keypoints (left shoulder, right shoulder, left hip, right hip) are available, and (2) there exists at least one sample in the index set whose 3D pose is less than $0.1$ NP-MPJPE apart. Due to the randomness in splitting query/index videos, we sample $5$ splits and report the averaged results for our experiments on this dataset.

\paragraph{MPII Human Pose (2DHP)}\cite{andriluka20142d} This dataset is commonly used for 2D pose estimation, containing about $25$K in-the-wild images. Since groundtruth 3D poses are not available, we only show qualitative results on this dataset.

\paragraph{Penn Action}\cite{zhang2013actemes} This dataset contains $2326$ trimmed videos for $15$ pose-based actions taken from $4$ different views. We follow the standard protocol~\cite{nie2015joint} for our action classification and video alignment experiments.

\subsection{Implementation Details} \label{sec:training_details}

For single-frame models, the backbone network architecture for our model is based on~\cite{martinez2017simple} for simplicity and fair comparison with 3D lifting. We use two residual FC blocks, batch normalization, $0.3$ dropout during training, and no weight norm clipping. Unless stated otherwise, we use embedding dimension $d=16$. To weigh different losses, we use $w_\text{ratio}=1$, $w_\text{positive}=0.005$, and $w_\text{prior}=0.001$. We choose $\beta=2$ for the triplet ratio loss margin and $K=20$ for the number of samples. During training, we normalize matching probabilities to within $[0.05,0.95]$ for numerical stability. The matching NP-MPJPE threshold is $\kappa=0.1$ for all training and evaluation. Ablation studies on hyperparameters can be found in Section~\ref{sec:ablation}. For Temporal Pr-VIPE, we embed 2D poses from $T=7$ frames. The temporal atrous sampling rate is chosen based on video frame rate, such that the input sequence covers approximately $0.5$ second. Specifically, we use atrous rate $4$ for $50$ FPS videos and $2$ for $25$ FPS videos.

We use PersonLab \cite{papandreou2018personlab} 2D keypoint detector for our experiments, unless stated otherwise, while our approach does not rely on a particular 2D keypoint detector. For random rotation during camera augmentation, we uniformly sample azimuth angle between $\pm180^{\circ}$, elevation angle between $\pm30^{\circ}$, and roll angle between $\pm30^{\circ}$. We use Adagrad optimizer~\cite{duchi2011adaptive} with fixed learning rate $0.02$, and batch size $N=256$. Our batch consists of an even mix of detected and projected 2D keypoints, which includes anchor and positive pairs from different sources. For the occlusion-robust model training, we evenly mix triplets with fully visible poses and triplets with partially visible anchor poses in a batch. Our implementation is in TensorFlow~\cite{tensorflow2015whitepaper}.

\begin{table*}[!t]
  \centering
\caption{Comparison of cross-view pose retrieval results Hit@$k$ ($\%$) on H3.6M and 3DHP. $^*$ indicates normalization and Procrustes alignment are required to compare query-index pairs.} \label{tab:3dhp}  
   \begin{tabular}{c | c c c c | c c c c | c c c c} 
   \toprule[0.2em]
   \multicolumn{1}{c|}{Dataset} & \multicolumn{4}{c|}{H3.6M} & \multicolumn{4}{c|}{3DHP (Chest)} & \multicolumn{4}{c}{3DHP (All)} \\
  $k$ & $1$ & $5$ & $10$ & $20$ & $1$ & $5$ & $10$ & $20$ & $1$ & $5$ & $10$ & $20$  \\
   \toprule[0.2em]
   2D keypoints* & $27.4$ & $41.4$ & $45.9$ & $49.7$ & $5.75$ &  $12.1$ & $12.1$ & $18.5$ & $10.3$ & $18.7$ & $22.6$ & $26.6$ \\
   3D lifting* & $68.8$ & $85.4$ & $89.5$ & $92.5$ & $26.0$ & $47.1$ & $55.9$ & $63.8$ & $25.5$ & $45.9$ & $54.6$ & $62.7$  \\
   $L2$-VIPE & $73.2$ & $90.4$ & $94.3$ & $96.9$ & $25.0$ & $48.2$ & $58.4$ & $68.1$ & $19.5$ & $38.6$ & $47.9$ & $57.4$ \\
   $L2$-VIPE (w/ aug.) & $72.4$ & $89.2$ & $93.2$ & $95.9$ & $\bm{29.1}$ & $52.7$ & $62.4$ & $71.3$ & $27.2$ & $49.9$ & $59.1$ & $68.1$  \\
   Pr-VIPE & $\bm{74.6}$ & $\bm{91.6}$ &  $\bm{95.2}$ & $\bm{97.5}$ & $25.7$ & $49.2$ & $59.8$ & $69.7$ & $19.8$ & $39.4$ & $49.1$ & $58.8$ \\
   Pr-VIPE (w/ aug.) & $72.9$ & $90.0$ & $93.9$ & $96.5$ & $29.0$ & $\bm{53.1}$ &  $\bm{63.0}$ & $\bm{72.0}$ & $\bm{27.3}$ & $\bm{50.0}$ & $\bm{59.9}$ & $\bm{69.0}$ \\
   \bottomrule[0.1em]
\end{tabular}
\end{table*}

\begin{table}
  \centering
  \caption{Comparison with recent methods of cross-view pose retrieval results Hit@$k$ ($\%$) on H3.6M hold-out subset. $^*$ indicates normalization and Procrustes alignment are required to compare query-index pairs. } \label{tab:new_retreival}
  \scalebox{1.0}{
   \begin{tabular}{c | c c c c c}
   \toprule[0.2em]
   & \multicolumn{4}{c}{H3.6M}\\
    Model & $k=1$ & $k=5$ & $k=10$ &  $k=20$  \\ [0.25ex]
   \toprule[0.2em]
  3D lifting* & $68.2$ & $85.5$ & $90.0$ & $93.5$ \\
  SemGCN* & $70.7$ & $88.4$ & $92.3$ & $95.1$ \\
  EpipolarPose* & $78.0$ & $92.2$ & $95.2$ & $97.1$ \\
  Pr-VIPE ($16$D) & $78.3$ & $94.2$ &  $97.2$ & $98.7$ \\
  Pr-VIPE ($32$D) & $\bm{81.6}$ & $\bm{95.5}$ &  $\bm{97.9}$ & $\bm{99.1}$ \\
   \bottomrule[0.1em]
\end{tabular}
}
\end{table}

\subsection{Cross-View Pose Retrieval}\label{sec:retrieval_subsection}

The goal of cross-view pose retrieval is to retrieve matching 2D poses from different camera views given a 2D pose from one view. This task is evaluated using multi-view human pose datasets, since we have access to 2D pose detections of the same 3D pose across views.

\subsubsection{Evaluation Procedure}\label{sec:full_body_procedure}
In this evaluation, we iterate through all camera pairs in the dataset. For each camera pair, we query using detected 2D keypoints from the first camera view and find the nearest neighbors in the embedding space from the second, different, camera view. Results averaged across all cameras pairs are reported.

We report Hit@$k$ with $k=1$, $5$, $10$, and $20$ on pose retrievals, which is the percentage of top-$k$ retrieved poses that have at least one accurate retrieval. A retrieval is considered accurate if the 3D groundtruth from the retrieved pose satisfies the matching function~(\ref{eq:12}) with $\kappa=0.1$.

On H3.6M and 3DHP, we compare Pr-VIPE with using 2D pose inputs, 2D-to-3D lifting models~\cite{martinez2017simple} and $L2$-VIPE. $L2$-VIPE outputs $L2$-normalized point embeddings, and is trained with the squared $L2$ distance kernel, similar to~\cite{schroff2015facenet}. For fair comparison, we use the same backbone network architecture for all the models. Notably, this architecture~\cite{martinez2017simple} has been tuned for lifting tasks on H3.6M. Since the estimated 3D poses in camera coordinates are not view-invariant, we apply normalization and Procrustes alignment to align the estimated 3D poses between index and query for retrieval. In comparison, our embeddings do not require any alignment or other post-processing during retrieval.

On H3.6M, we further compare with two recent 3D pose estimation methods: Semantic Graph Convolutional Networks (SemGCN)~\cite{zhao2019semantic} and EpipolarPose~\cite{kocabas2019self}. 
SemGCN leverages graph convolutional networks~\cite{kipf2017semi} to encode 2D poses with a skeleton-based graph representation. We train SemGCN on 2D poses from a Cascaded Pyramid Network (CPN)~\cite{chen2018cascaded} pose detector, which takes bounding boxes detected by Mask R-CNN~\cite{he2017mask} as inputs. Following~\cite{pavllo20193d}, the CPN model is pre-trained on COCO~\cite{lin2014microsoft} and then fine-tuned on H3.6M. EpipolarPose is an image-based 3D pose estimation model. We report the results from its fully supervised version. 

For Pr-VIPE, we retrieve poses using nearest neighbors in the embedding space with respect to the sampled matching probability~(\ref{eq:9}), which we refer to as retrieval confidence later in this paper.

\subsubsection{Quantitative Results}

Table~\ref{tab:3dhp} shows the comparison between Pr-VIPE and baseline methods on the H3.6M hold-out set and 3DHP. The H3.6M hold-out set has the same camera views as the training set, and 3DHP has rich novel views and poses unseen from the H3.6M training set. When we use all cameras from 3DHP, we evaluate model generalization to both novel poses and views. When we evaluate using only the $5$ chest-level cameras from 3DHP, the views are more similar to the training set from H3.6M. This enables us to focus more on evaluating model generalization to novel poses.

Without camera augmentation, Pr-VIPE is able to generally perform better than the other baselines on H3.6M and 3DHP (chest-level cameras). This observation indicates that Pr-VIPE is able to generalize as well as other baseline methods to novel poses. Using camera augmentation significantly improves the performance of Pr-VIPE on 3DHP for both chest-level cameras and all cameras. This observation indicates that camera augmentation improves model generalization to novel views. The same results can be observed for $L2$-VIPE for chest-level and all cameras. We note that $L2$-VIPE and Pr-VIPE has similar performance on 3DHP with camera augmentation, while Pr-VIPE outperforms $L2$-VIPE on H3.6M under this setting and performs consistently better on both H3.6M and 3DHP without camera augmentation. On H3.6M, camera augmentation slightly reduces accuracy for both Pr-VIPE and $L2$-VIPE, which is likely because it reduces overfitting to the training camera views.
In general, by performing camera augmentation, which does not require camera parameters or additional groundtruth information, Pr-VIPE is able to generalize better to novel poses and views.

Additionally, we note that 3D lifting models can generalize relatively well to novel views with the help of the additional Procrustes alignment, which requires expensive singular value decomposition computation between every index-query pair. We applied similar camera augmentation to the lifting model, but did not see improvements in performance. It is also worth mentioning that the first row in Table~\ref{tab:3dhp} shows retrieval accuracy using aligned 2D pose keypoints. The poor retrieval performance confirms the fact that models must learn view invariance to perform well on this task.

Table~\ref{tab:new_retreival} contains comparisons of Pr-VIPE to additional more recent 3D pose estimation baselines. For this comparison, we follow the standard protocol used in~\cite{kocabas2019self,zhao2019semantic} and evaluate on every $64$th frame of the H3.6M hold-out set. Table~\ref{tab:new_retreival} demonstrates that Pr-VIPE, without the need for Procrustes alignment, is able to achieve higher retrieval accuracy compared with 3D pose estimation models. In addition, we are able to further improve performance using higher dimensional Pr-VIPE, without the need for additional training data.

\subsubsection{Qualitative Results}

\def\figsize{0.15
}
\def\fighspace{-2mm}
\def\fighspacer{+1.5mm}
\begin{figure*}
\centering
\begin{tabular}{cccccc}
\centering
\scriptsize{$C=0.993$}\hspace{\fighspace} & \scriptsize{$E=0.04$}\hspace{\fighspacer} & \scriptsize{$C=0.968$}\hspace{\fighspace} & \scriptsize{$E=0.07$}\hspace{\fighspacer} & \scriptsize{$C=0.714$}\hspace{\fighspace} & \scriptsize{$E=0.06$}\hspace{\fighspacer} \\
\includegraphics[width=\figsize\textwidth]{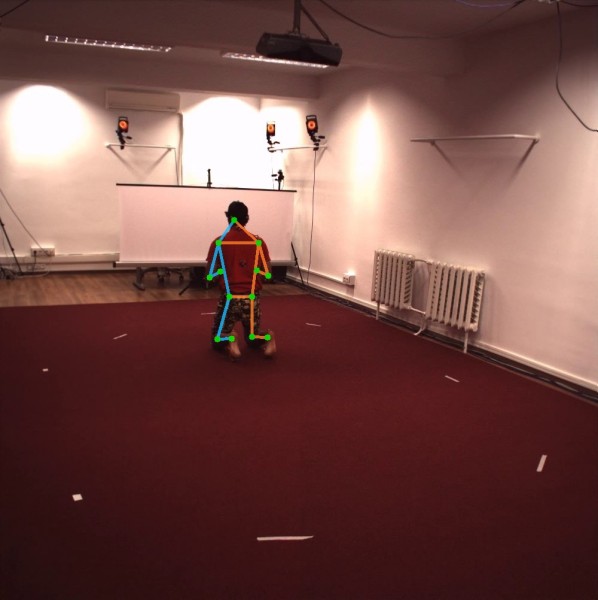}\hspace{\fighspace} & \includegraphics[width=\figsize\textwidth]{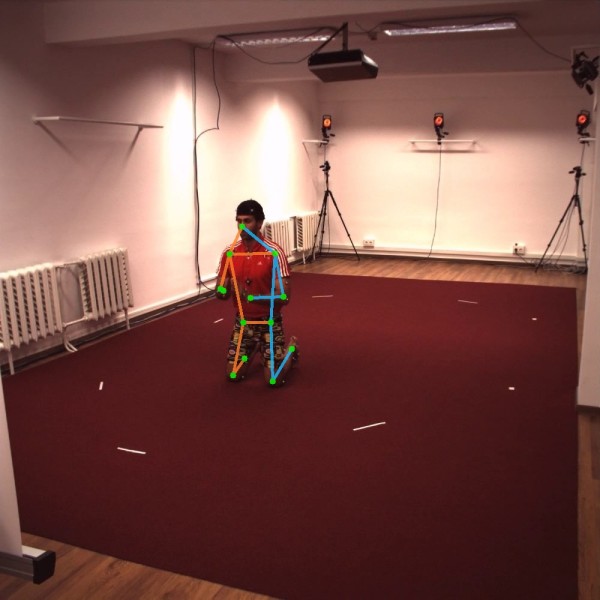}\hspace{\fighspacer}  & \includegraphics[width=\figsize\textwidth]{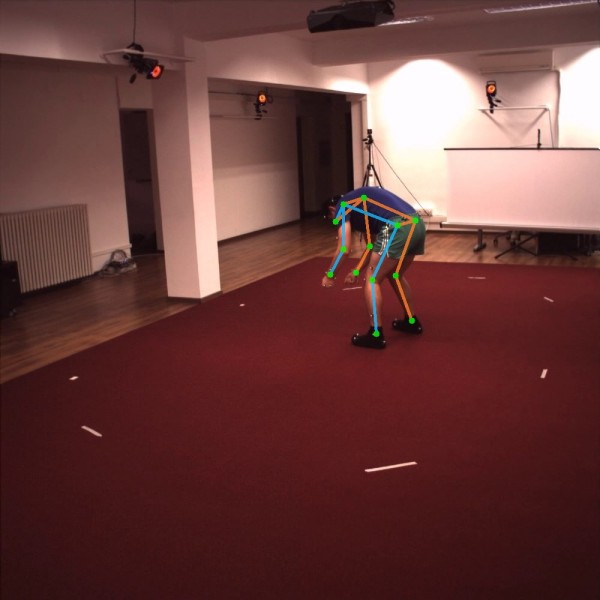}\hspace{\fighspace} & \includegraphics[width=\figsize\textwidth]{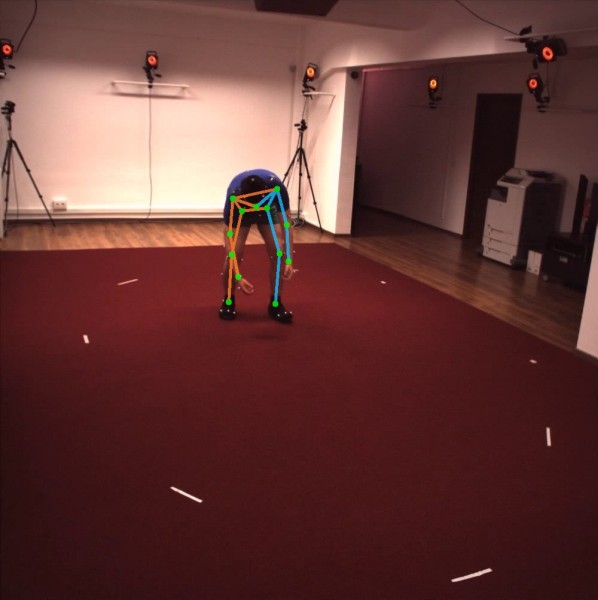}\hspace{\fighspacer} &
\includegraphics[width=\figsize\textwidth]{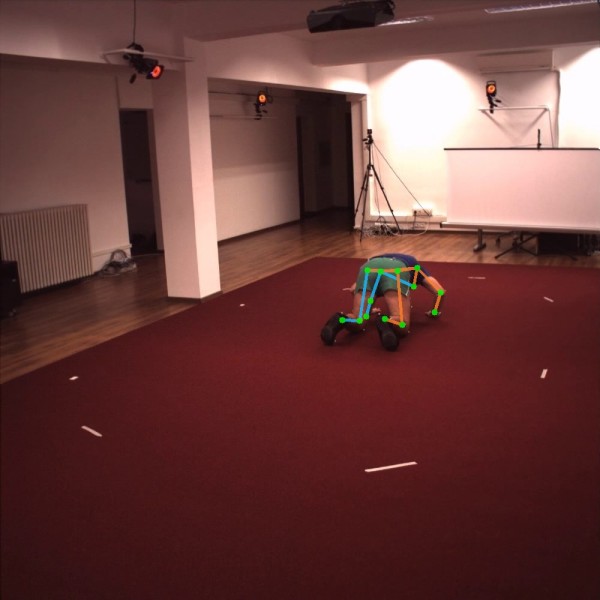}\hspace{\fighspace} & \includegraphics[width=\figsize\textwidth]{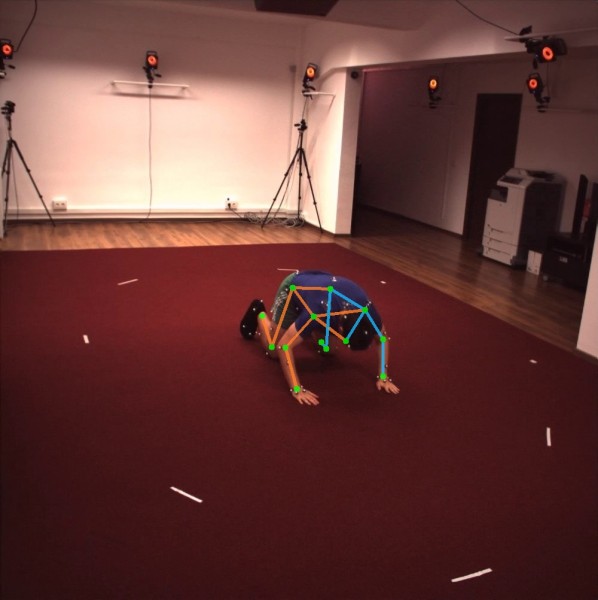}\hspace{\fighspacer} \\

\scriptsize{$C=0.836$}\hspace{\fighspace} & \scriptsize{$E=0.00$}\hspace{\fighspacer} & \scriptsize{$C=0.949$}\hspace{\fighspace} & \scriptsize{$E=0.13$}\hspace{\fighspacer} & \scriptsize{$C=0.926$}\hspace{\fighspace} & \scriptsize{$E=0.10$}\hspace{\fighspacer} \\
\includegraphics[width=\figsize\textwidth]{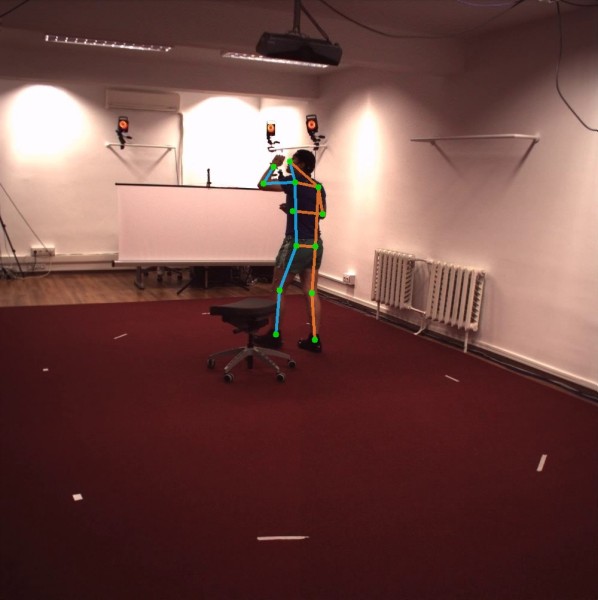}\hspace{\fighspace} & \includegraphics[width=\figsize\textwidth]{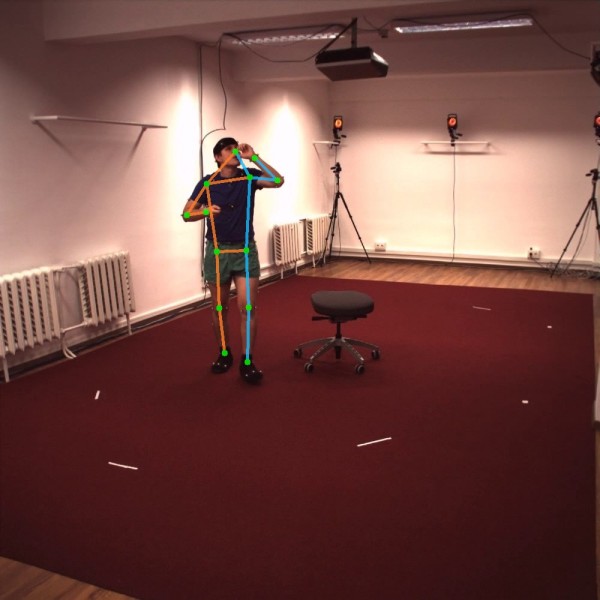}\hspace{\fighspacer}  & \includegraphics[width=\figsize\textwidth]{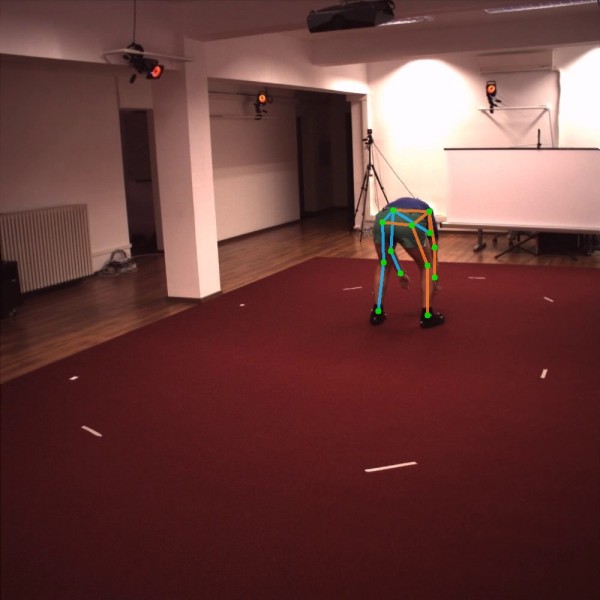}\hspace{\fighspace} & \includegraphics[width=\figsize\textwidth]{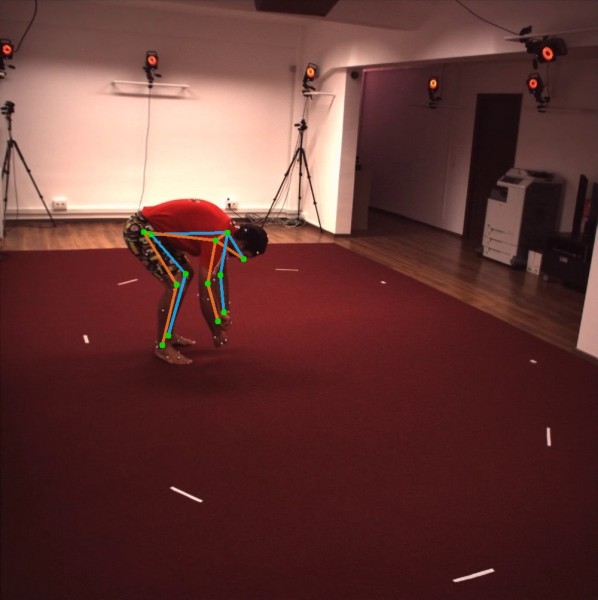}\hspace{\fighspacer} &
\includegraphics[width=\figsize\textwidth]{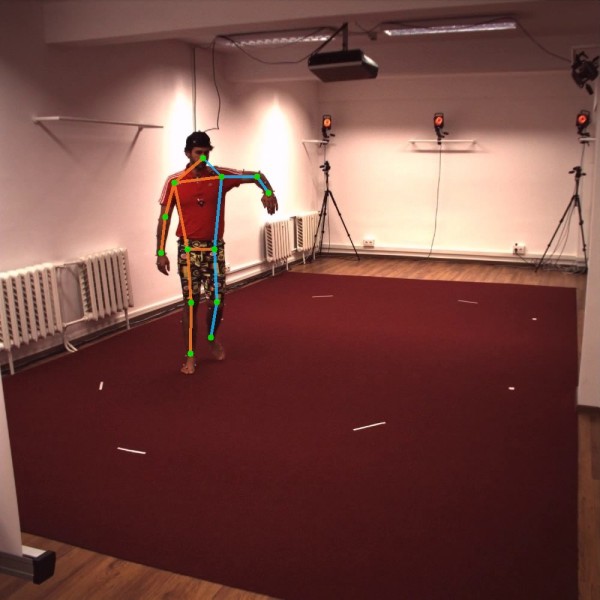}\hspace{\fighspace} & \includegraphics[width=\figsize\textwidth]{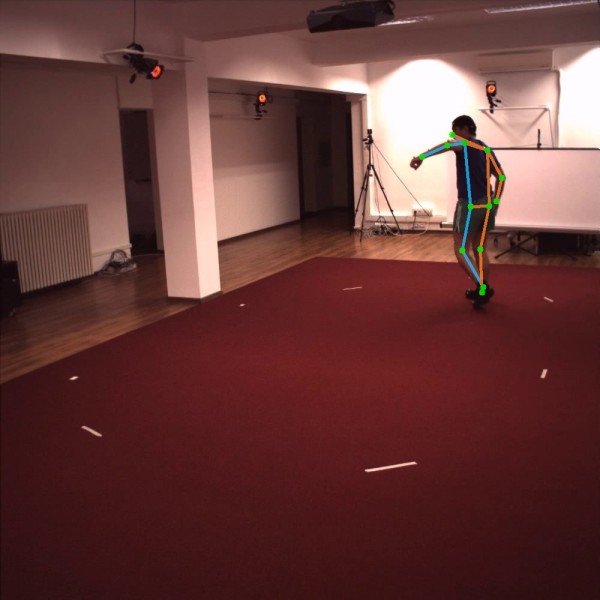}\hspace{\fighspacer} \\

\scriptsize{$C=0.969$}\hspace{\fighspace} & \scriptsize{$E=0.00$}\hspace{\fighspacer} & \scriptsize{$C=0.961$}\hspace{\fighspace} & \scriptsize{$E=0.06$}\hspace{\fighspacer} & \scriptsize{$C=0.606$}\hspace{\fighspace} & \scriptsize{$E=0.00$}\hspace{\fighspacer} \\
\includegraphics[width=\figsize\textwidth]{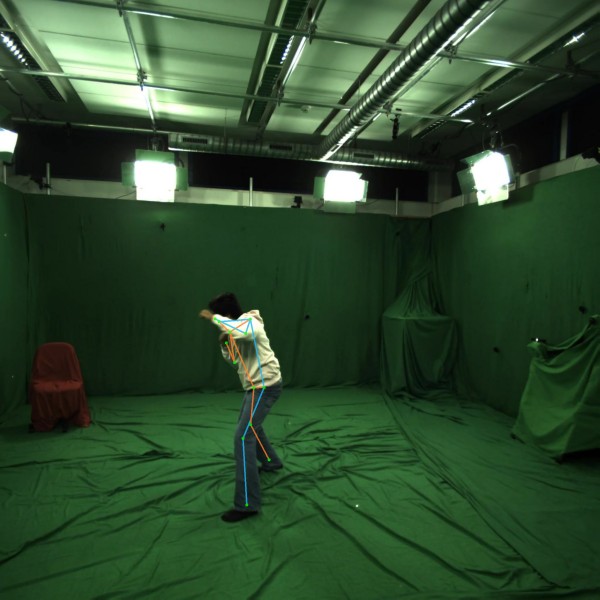}\hspace{\fighspace} & \includegraphics[width=\figsize\textwidth]{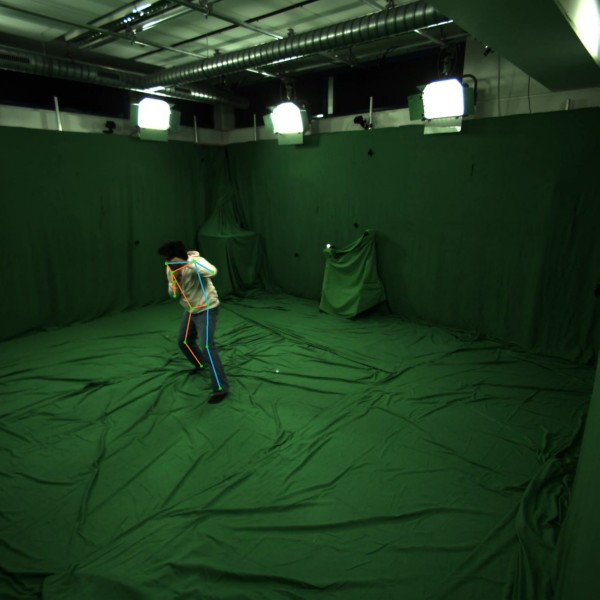}\hspace{\fighspacer}  & \includegraphics[width=\figsize\textwidth]{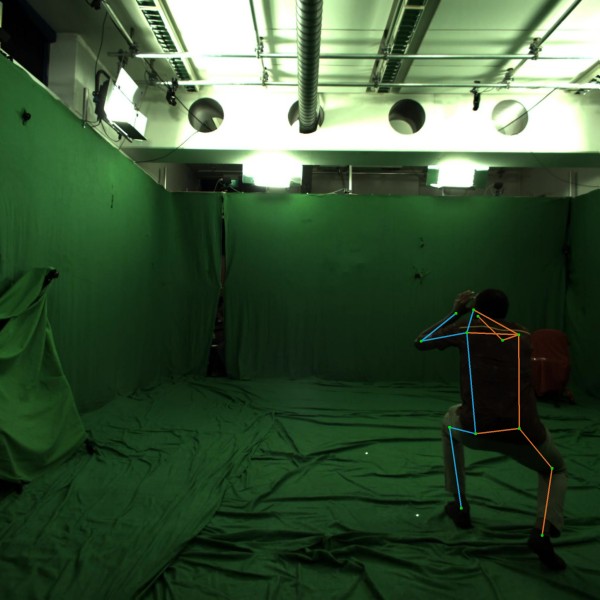}\hspace{\fighspace} & \includegraphics[width=\figsize\textwidth]{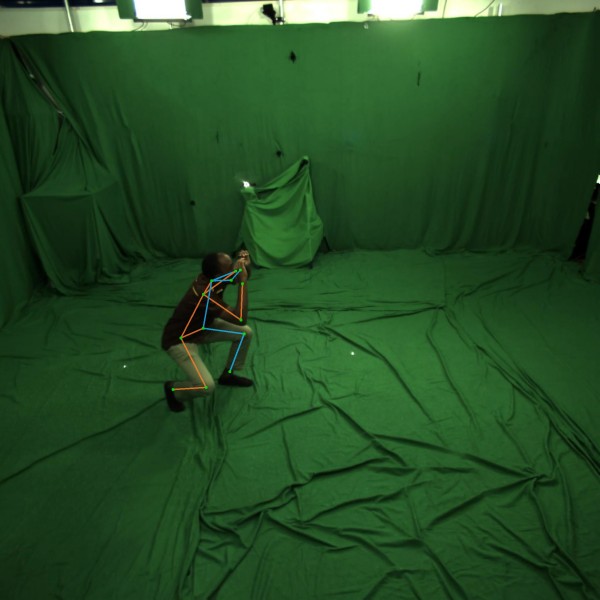}\hspace{\fighspacer} &
\includegraphics[width=\figsize\textwidth]{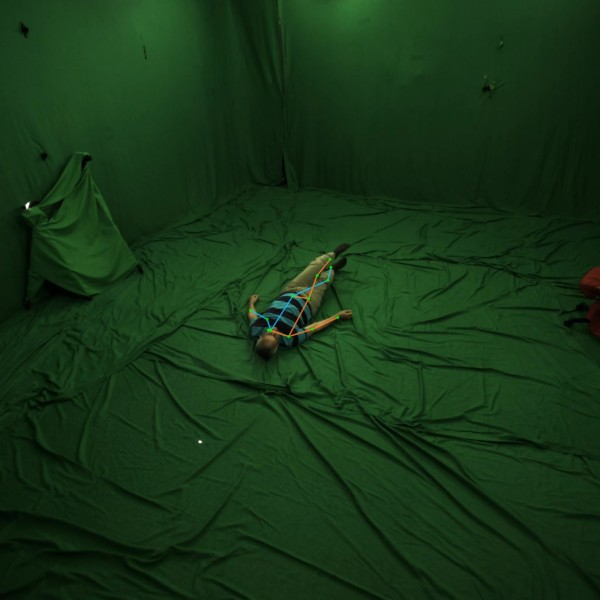}\hspace{\fighspace} & \includegraphics[width=\figsize\textwidth]{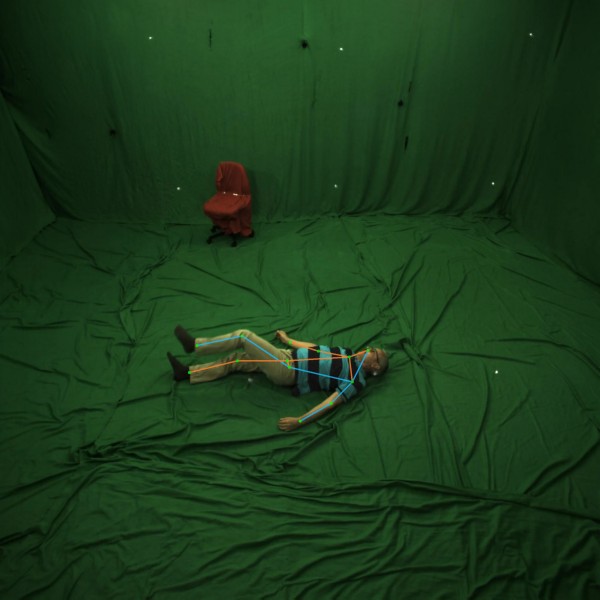}\hspace{\fighspacer} \\

\scriptsize{$C=0.828$}\hspace{\fighspace} & \scriptsize{$E=0.16$}\hspace{\fighspacer} & \scriptsize{$C=0.946$}\hspace{\fighspace} & \scriptsize{$E=0.06$}\hspace{\fighspacer} & \scriptsize{$C=0.889$}\hspace{\fighspace} & \scriptsize{$E=0.32$}\hspace{\fighspacer} \\
\includegraphics[width=\figsize\textwidth]{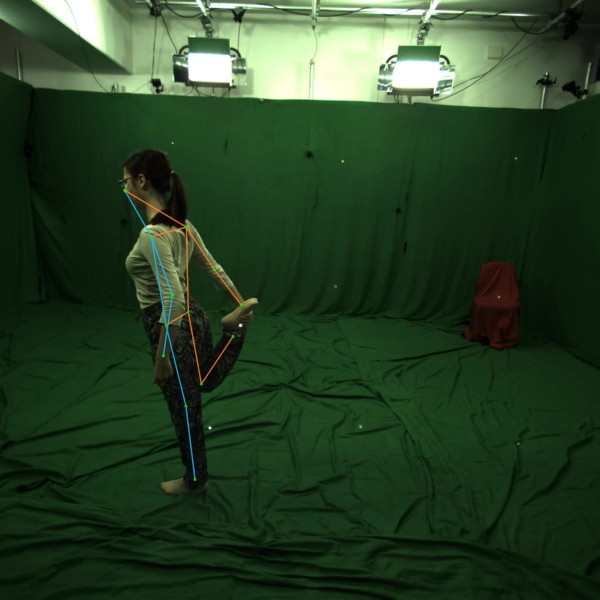}\hspace{\fighspace} & \includegraphics[width=\figsize\textwidth]{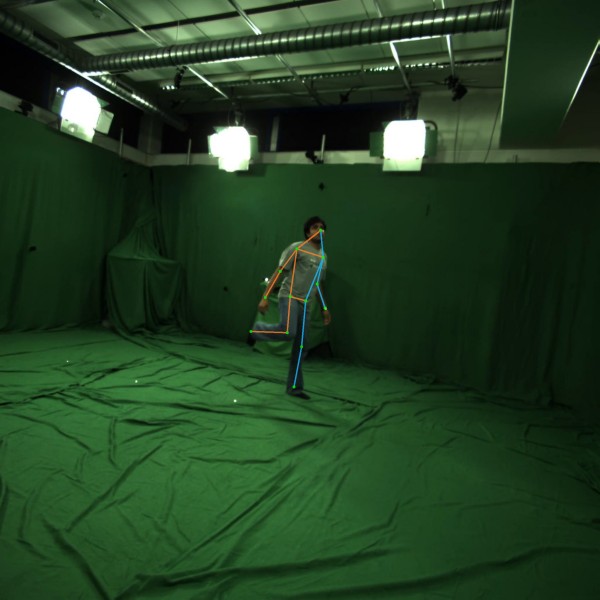}\hspace{\fighspacer}  & \includegraphics[width=\figsize\textwidth]{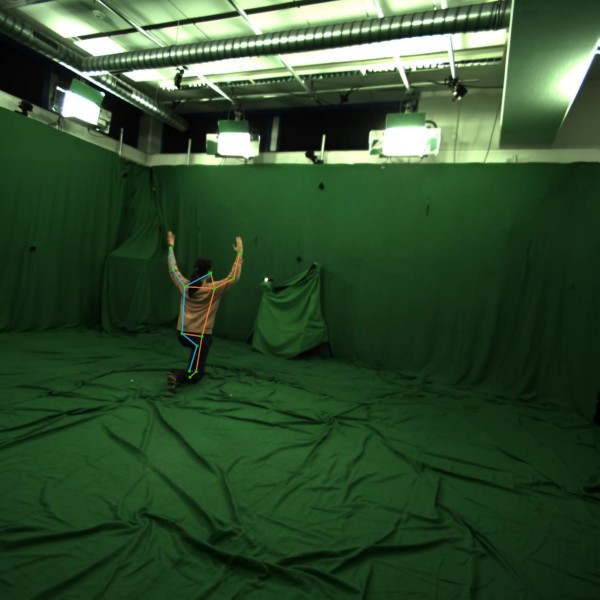}\hspace{\fighspace} & \includegraphics[width=\figsize\textwidth]{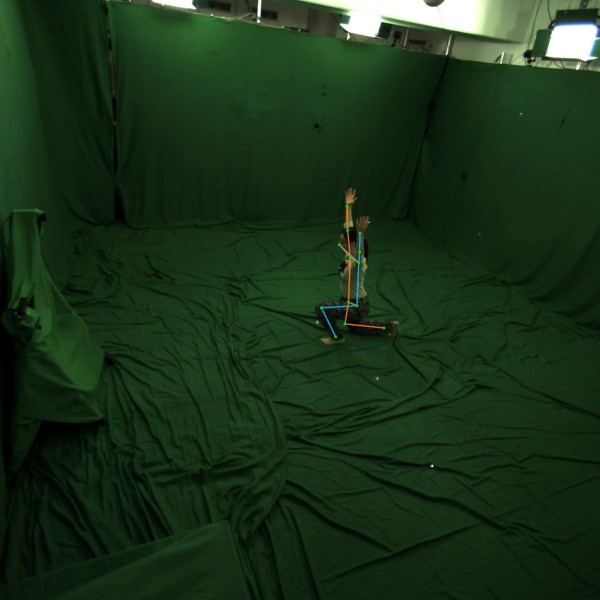}\hspace{\fighspacer} &
\includegraphics[width=\figsize\textwidth]{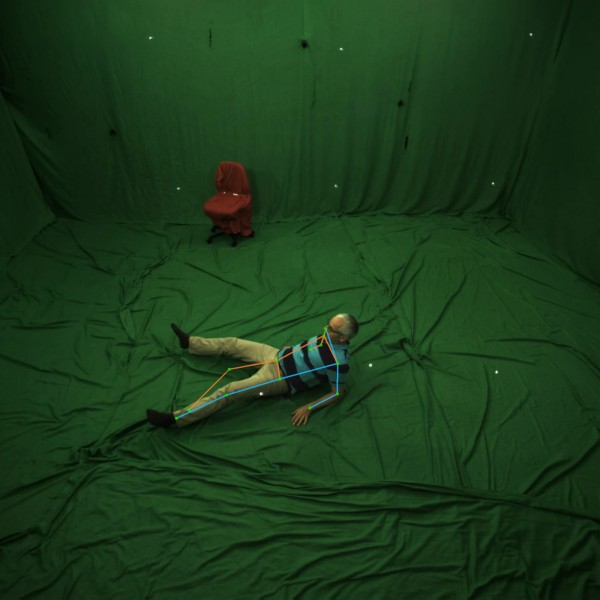}\hspace{\fighspace} & \includegraphics[width=\figsize\textwidth]{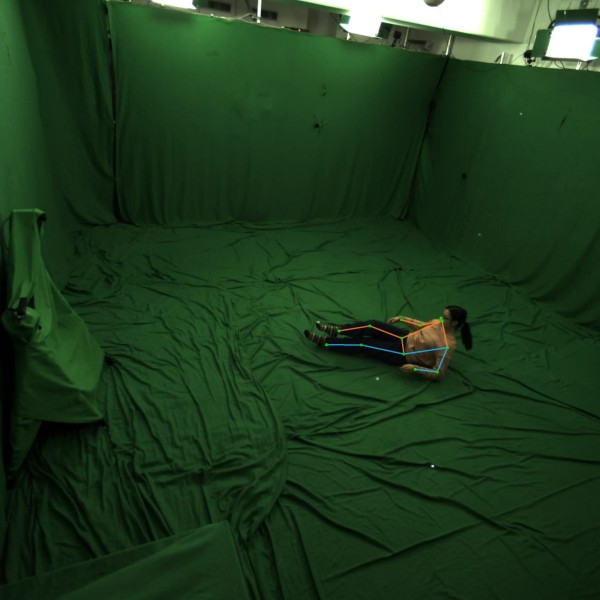}\hspace{\fighspacer} \\

\scriptsize{$C=0.729$}\hspace{\fighspace} & \hspace{\fighspacer} & \scriptsize{$C=0.777$}\hspace{\fighspace} & \hspace{\fighspacer} & \scriptsize{$C=0.737$}\hspace{\fighspace} & \hspace{\fighspacer} \\
\includegraphics[width=\figsize\textwidth]{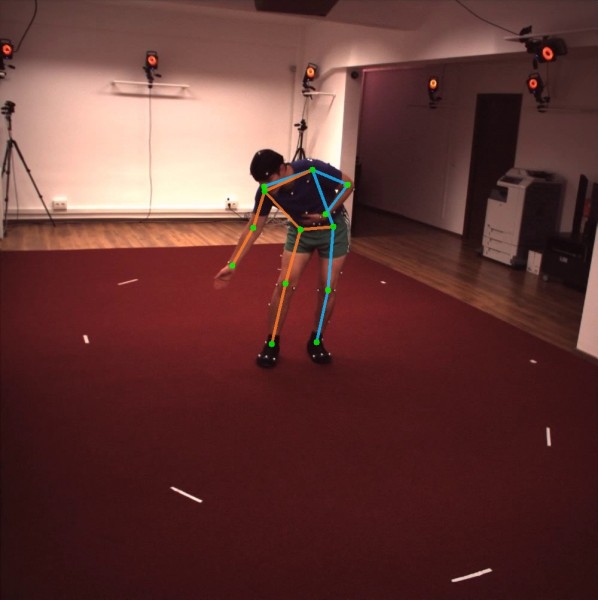}\hspace{\fighspace} & \includegraphics[width=\figsize\textwidth]{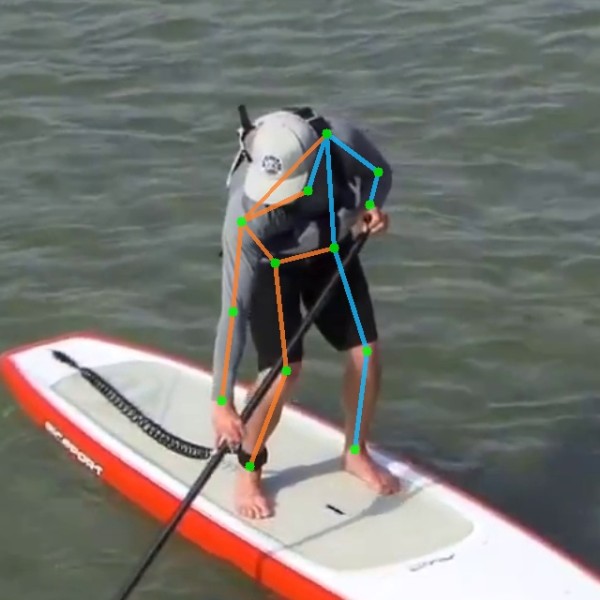}\hspace{\fighspacer}  & \includegraphics[width=\figsize\textwidth]{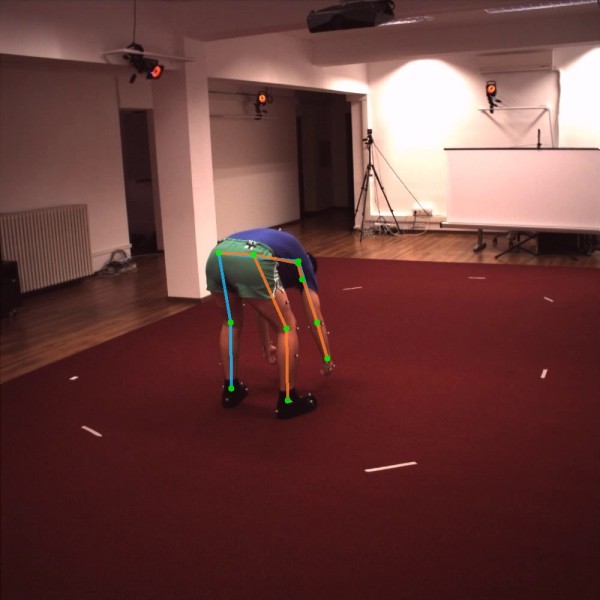}\hspace{\fighspace} & \includegraphics[width=\figsize\textwidth]{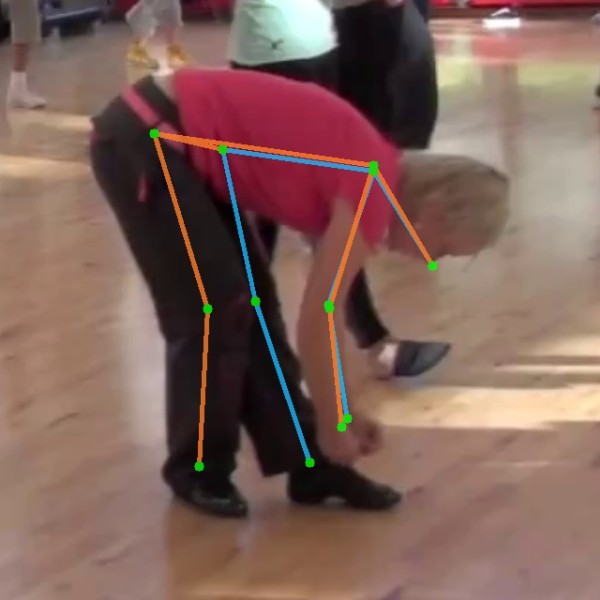}\hspace{\fighspacer} &
\includegraphics[width=\figsize\textwidth]{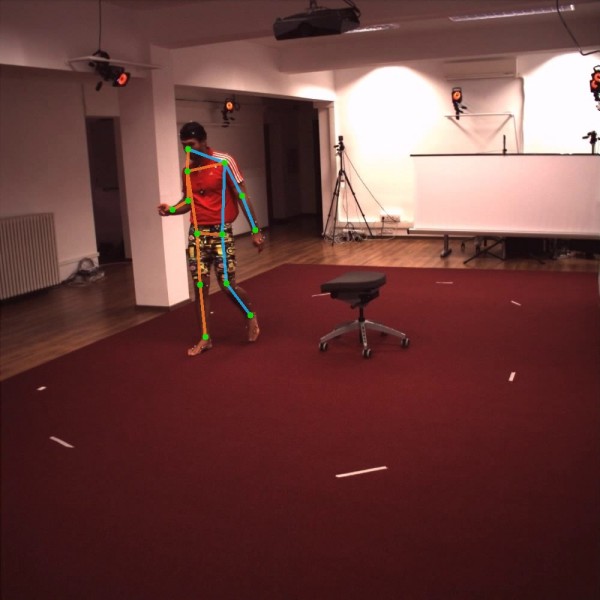}\hspace{\fighspace} & \includegraphics[width=\figsize\textwidth]{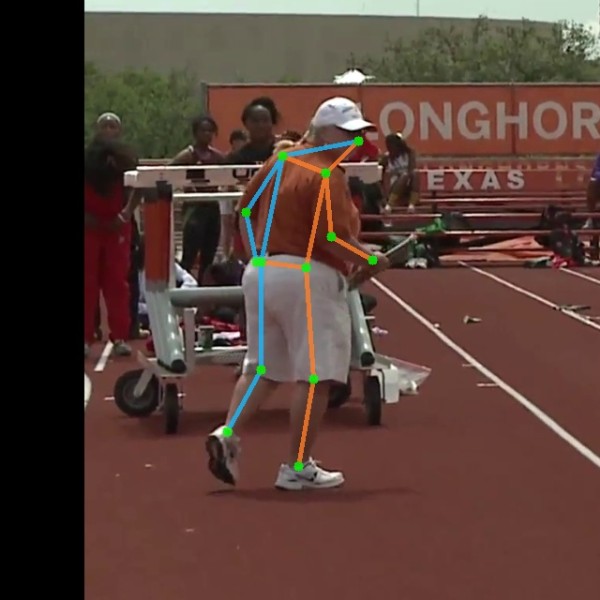}\hspace{\fighspacer} \\

\scriptsize{$C=0.625$}\hspace{\fighspace} & \hspace{\fighspacer} & \scriptsize{$C=0.636$}\hspace{\fighspace} & \hspace{\fighspacer} & \scriptsize{$C=0.760$}\hspace{\fighspace} & \hspace{\fighspacer} \\
\includegraphics[width=\figsize\textwidth]{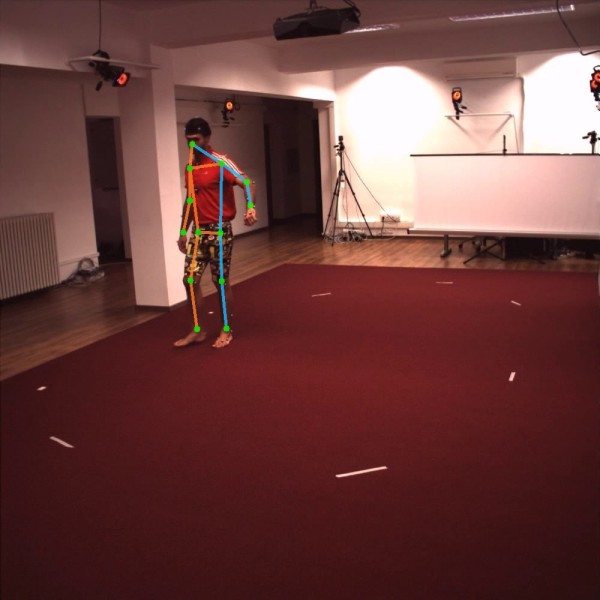}\hspace{\fighspace} & \includegraphics[width=\figsize\textwidth]{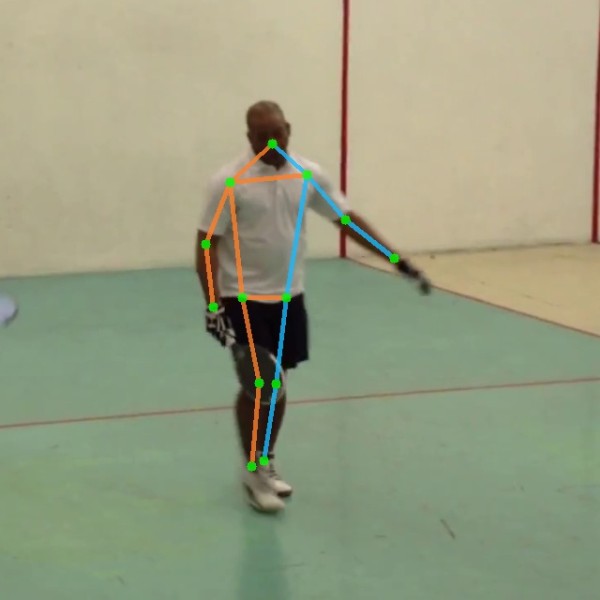}\hspace{\fighspacer}  & \includegraphics[width=\figsize\textwidth]{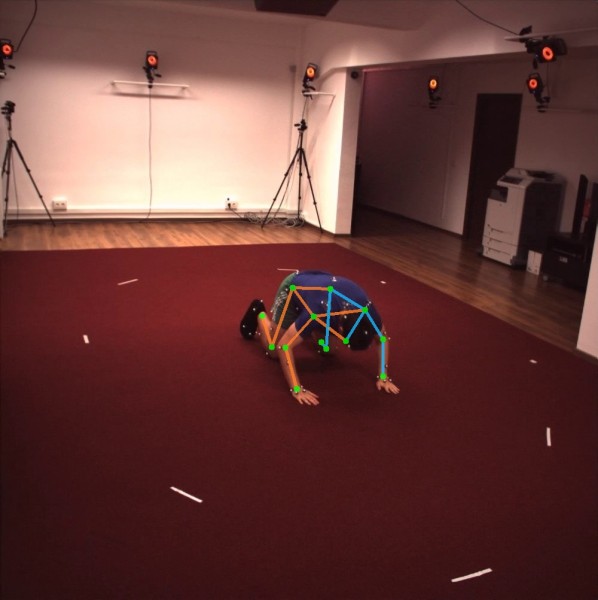}\hspace{\fighspace} & \includegraphics[width=\figsize\textwidth]{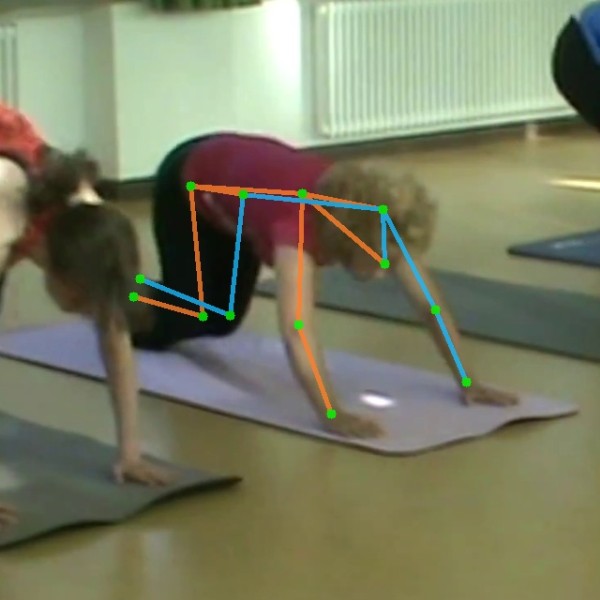}\hspace{\fighspacer} &
\includegraphics[width=\figsize\textwidth]{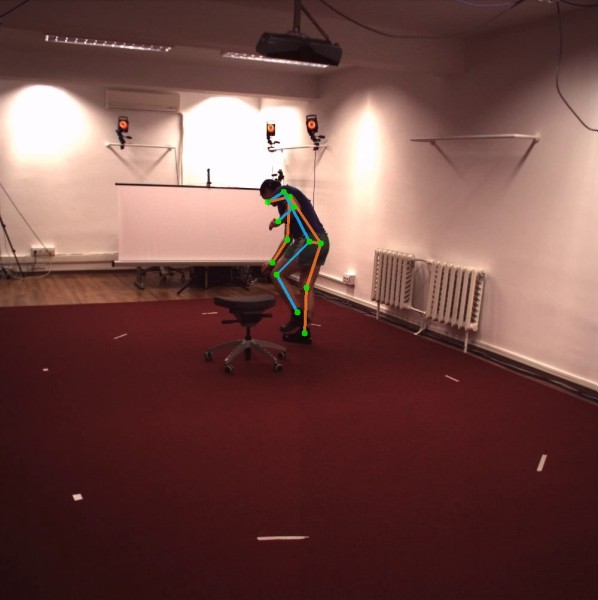}\hspace{\fighspace} & \includegraphics[width=\figsize\textwidth]{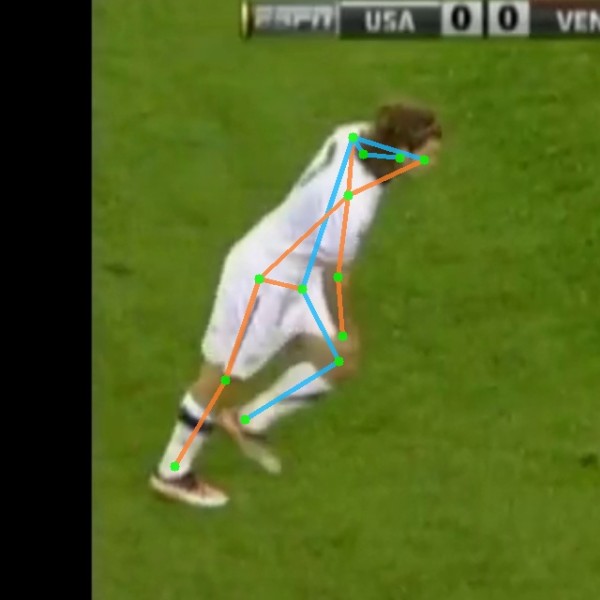}\hspace{\fighspacer} \\

\scriptsize{$C=0.839$}\hspace{\fighspace} & \hspace{\fighspacer} & \scriptsize{$C=0.890$}\hspace{\fighspace} & \hspace{\fighspacer} & \scriptsize{$C=0.944$}\hspace{\fighspace} & \hspace{\fighspacer} \\
\includegraphics[width=\figsize\textwidth]{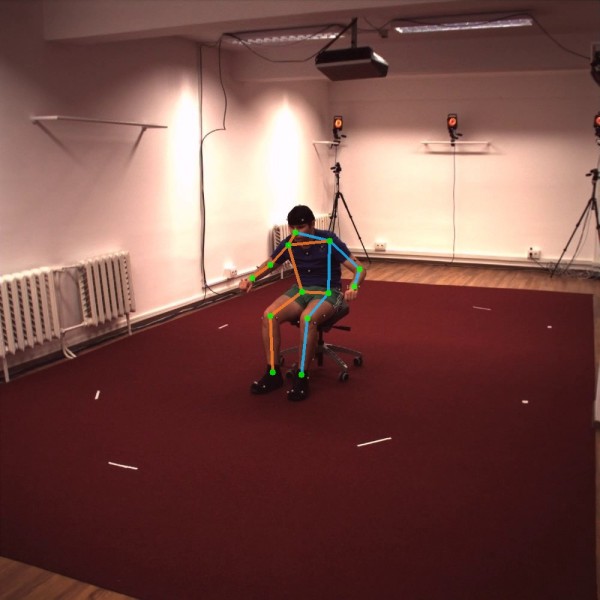}\hspace{\fighspace} & \includegraphics[width=\figsize\textwidth]{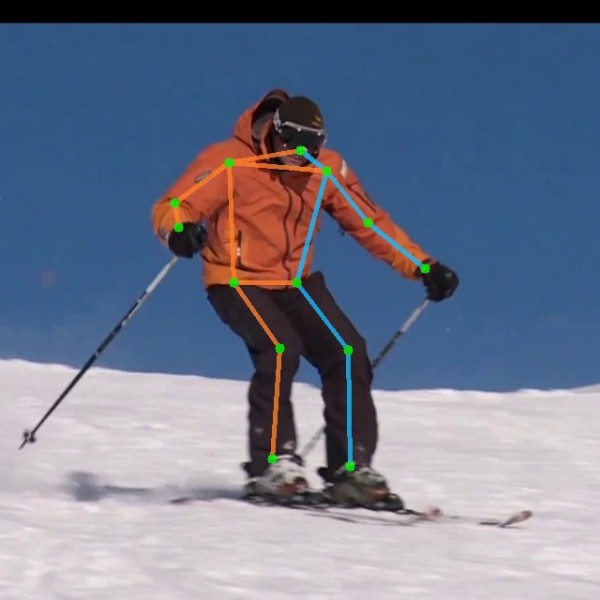}\hspace{\fighspacer}  & \includegraphics[width=\figsize\textwidth]{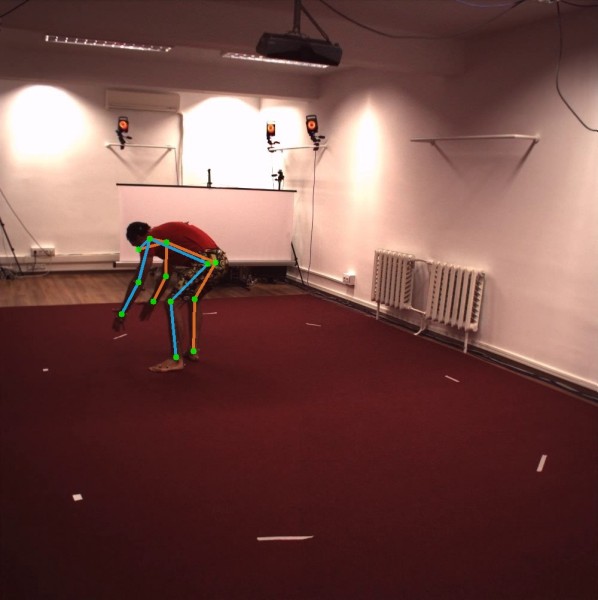}\hspace{\fighspace} & \includegraphics[width=\figsize\textwidth]{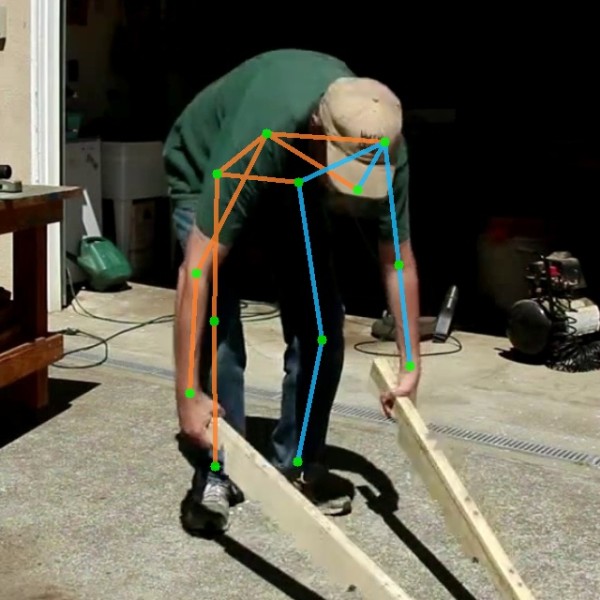} \hspace{\fighspacer} &
\includegraphics[width=\figsize\textwidth]{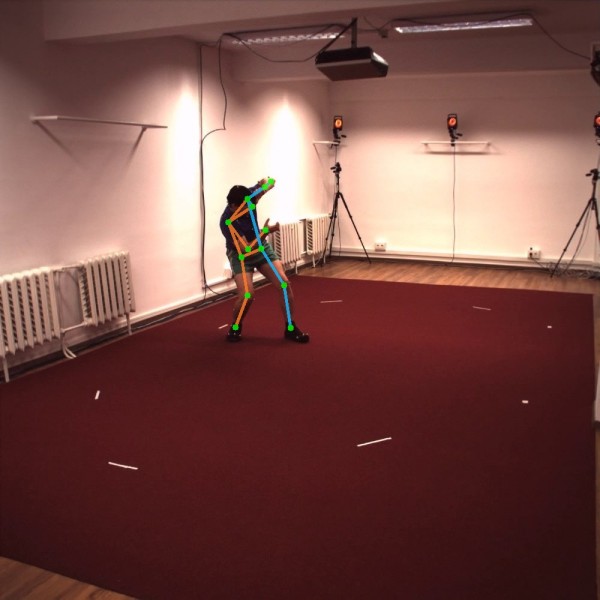}\hspace{\fighspace} & \includegraphics[width=\figsize\textwidth]{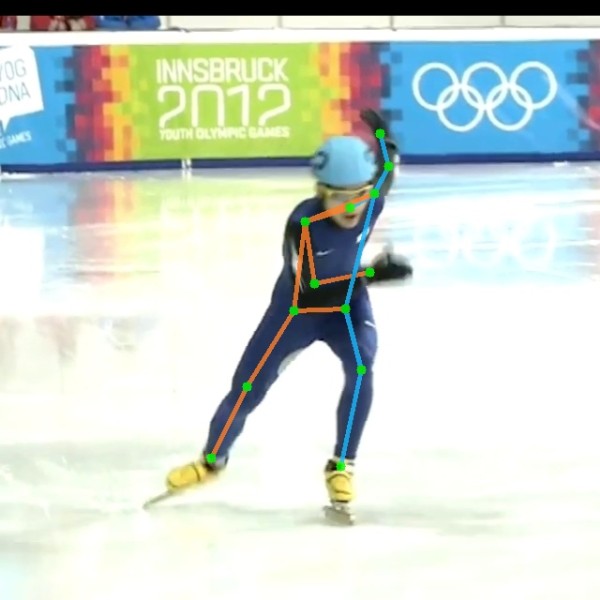}\hspace{\fighspacer} \\

\end{tabular}
\caption{Visualization of pose retrieval results. The first and second row are from H3.6M; the third and fourth row are from 3DHP; the last three rows are using queries from H3.6M to retrieve from 2DHP. On each row, we show the query pose on the left for each image pair and the top-$1$ retrieval using Pr-VIPE (with camera augmentation) on the right. We also show the retrieval confidences (``$C$'') and the top-$1$ NP-MPJPEs (``$E$'', if 3D pose groundtruth is available).}
\label{fig:retrieval}
\end{figure*}

Fig.~\ref{fig:retrieval} shows qualitative retrieval results using Pr-VIPE (with camera augmentation). 
As shown in the first and second rows, the retrieval confidences are generally high for H3.6M. This indicates that the retrieved poses are close to their queries in the embedding space.
In the third and fourth rows, the retrieval confidences are generally lower for 3DHP compared with H3.6M. 
This is likely because there are novel poses and views unseen during training, which results in the nearest neighbor being slightly farther away in the embedding space. 
We see that the model can generalize to novel views as the 3DHP images are taken at different camera elevations from H3.6M.

We show qualitative results using queries from the H3.6M hold-out set to retrieve from 2DHP in the last three rows of Fig.~\ref{fig:retrieval}. These results demonstrate that as long as the 2D keypoint detections are reasonably reliable, our model is able to retrieve poses across views and subjects and works on in-the-wild images, despite being trained only with in-lab data (H3.6M).

On the other hand, these examples also indicate our model relies on the quality of 2D keypoint input as a limitation to our approach. Errors in 2D keypoint detection can lead to retrieval errors as shown by the rightmost pair on row 4 (Fig.~\ref{fig:retrieval}). This example has a large retrieval error due to an erroneous right leg keypoint detection in the query pose. 
Additionally, projection ambiguity can also cause incorrect retrievals. In the rightmost pair on row 7, the query pose has the left arm in front of the body, and the retrieved pose has the left arm in the back. This is likely caused by the ambiguous appearance of the 2D pose skeleton without image context.

\subsection{Cross-View Pose Sequence Retrieval}\label{sec:sequence_retrieval}
In this section, we evaluate our Temporal Pr-VIPE models for cross-view pose sequence retrieval. Similar to cross-view pose retrieval, this task targets evaluating temporal embedding quality in terms of retrieving pose sequences to match a query sequence from a different camera view.

\subsubsection{Evaluation Procedure}
Given two pose sequences of the same length, we first compute the NP-MPJPE between each corresponding pose pair from both sequences. Then the maximum of these pairwise NP-MPJPEs is defined as the NP-MPJPE between the two sequences. We choose maximum here to reflect our requirement for the two sequences to be strictly close at all timestamps.

For each query sequence, we retrieve its $k$ nearest neighbor sequences from an index set based on their embedding distances. Then the sequence NP-MPJPE between each query and retrieval pair is thresholded to determine whether each retrieval is a correct match. Similar to the single-frame pose retrieval task, we evaluate on the H3.6M and the 3DHP dataset. We iterate through all camera pairs in each dataset as query and index, and report averaged results across all such camera pairs. 

We compare our Temporal Pr-VIPE result with baseline methods that stack single-frame embeddings within the same frame window into higher dimensional embeddings for retrieval distance computation. As we shall show in Section~\ref{sec:sequence_retrieval:quantitative_results}, simple stacking is an effective way of combining frame-level embeddings for sequence retrieval. However, it comes with a major drawback of high embedding dimensions, which can be prohibitive for large-scale applications. In this experiment, we demonstrate that with Temporal Pr-VIPE, we are able to achieve competitive retrieval performance with a much smaller embedding dimension.

\subsubsection{Quantitative Results}\label{sec:sequence_retrieval:quantitative_results}
From Table~\ref{tab:pose_sequence_retrieval_results}, we see that using Temporal Pr-VIPE, we are able to achieve competitive results with a much lower embedding dimension. Specifically, our $32$D temporal Pr-VIPE outperforms stacking $7$ $8$D single-frame embeddings (total $56$D) by a large margin. It also achieves slightly better performance compared to stacking $7$ $16$D embeddings (total $112$D), with fewer than one third of the embedding dimensions. We observe a similar trend across both H3.6M and 3DHP (unseen test set with new poses and new views), suggesting that both models have similar generalization abilities to new poses and new views. When we vary the number of dimensions for temporal Pr-VIPE, we note that at least $32$D embeddings is needed to achieve comparable performance to stacking $16$D Pr-VIPE (total $112$D), and at least $56$D embeddings is needed to achieve comparable performance to stacking $32$D Pr-VIPE (total $224$D). Additionally, we note that when the output dimensions are comparable, at total $56$D, Temporal Pr-VIPE performs much better than stacking $7$ $8$D Pr-VIPE.

\begin{table*}[!t]
  \centering
\caption{Comparison of cross-view pose sequence retrieval results Hit@$k$ ($\%$) on H3.6M and 3DHP. All the models in the table use camera augmentation.} \label{tab:pose_sequence_retrieval_results}  
   \begin{tabular}{c | c | c c c c | c c c c | c c c c} 
   \toprule[0.2em]
   \multicolumn{1}{c|}{Dataset} & \multirow{2}{*}{Total dim.} & \multicolumn{4}{c|}{H3.6M} & \multicolumn{4}{c|}{3DHP (Chest)} & \multicolumn{4}{c}{3DHP (All)} \\
  $k$ & & $1$ & $5$ & $10$ & $20$ & $1$ & $5$ & $10$ & $20$ & $1$ & $5$ & $10$ & $20$  \\
   \toprule[0.2em]
   Pr-VIPE (stacking $8$D) & $56$ & $70.3$ & $85.4$ & $89.8$ & $93.1$ & $41.9$ & $60.6$ & $67.8$ & $74.7$ & $38.9$ & $56.9$ & $64.3$ & $71.3$  \\
   Pr-VIPE (stacking $16$D) & $112$ & $78.7$ & $91.1$ & $94.3$ & $96.6$ & $48.9$ & $67.4$ & $74.3$ & $80.5$ & $45.0$ & $63.2$ & $70.2$ & $76.8$  \\
    Pr-VIPE (stacking $32$D) & $224$ & $80.8$ & $92.8$  & $95.8$ & $97.7$ & $50.4$ & $69.3$ & $76.2$ & $82.4$ & $47.0$ & $65.4$ & $72.4$ & $78.9$\\   
    \hline 
   Temporal Pr-VIPE ($16$D) & $16$ & $72.2$ & $87.9$ & $91.9$ & $94.9$ & $39.5$ & $60.8$ & $68.7$ & $75.7$ & $36.8$ & $57.8$ & $65.9$ & $73.2$\\
   Temporal Pr-VIPE ($32$D) & $32$ & $80.0$ & $92.1$ & $95.0$ & $97.1$ & $49.2$ & $69.1$ & $75.7$ & $81.3$ & $46.7$ & $66.4$ & $73.3$ & $79.3$  \\
   Temporal Pr-VIPE ($56$D) & $56$ & $80.4$ & $92.3$ & $95.1$ & $97.1$ & $50.4$ & $70.1$ & $76.4$ & $82.0$ & $47.7$ & $67.2$ & $73.9$ & $79.8$\\   
   \bottomrule[0.1em]
\end{tabular}
\end{table*}

\subsection{Partially-Visible Pose Retrieval}\label{sec:partial_retrieval}

In this section, we evaluate our occlusion-robust embedding for cross-view retrieval of partial 2D poses on H3.6M and 3DPW. We use the outdoor 3DPW dataset to test our embedding performance on realistic occlusions. Since there are limited realistic occlusions in H3.6M, we synthetically occlude certain keypoints and create two types of test sets for evaluation, namely targeted occlusion and real-distribution occlusion. The targeted occlusion set consists of $10$ hand-picked visibility patterns and aims at testing model performance in targeted partial pose search. The real-distribution occlusion consists of the top-$50$ most common visibility patterns in the wild and aims at testing models' potential performance on random photos in the wild. More details are provided in Section~\ref{sec:partial_vis_pose_retrieval_eval_proc}. For all the experiments in this section, we adopt the Pr-VIPE with camera augmentation setting (Section~\ref{sec:pr_vipe}), and apply different keypoint occlusion augmentation strategies during training (Section~\ref{sec:keypoint_availability_aug}).

\subsubsection{Evaluation Procedure}\label{sec:partial_vis_pose_retrieval_eval_proc}

Our evaluation procedure is similar to the full body retrieval procedure above in Section~\ref{sec:full_body_procedure}. We report retrieval performance with Hit@$k$ with $k=1$, $5$, $10$, and $20$, and a retrieval is considered accurate if the retrieved 3D groundtruth pose satisfies the matching function~(\ref{eq:12}) with $\kappa=0.1$ in terms of visible keypoints.

On H3.6M, we conduct two types of synthetic tests for model performance with different keypoint visibility patterns: 

\paragraph{Targeted Occlusion} We define $10$ visibility patterns for the targeted occlusions. These patterns are: missing left or right or both arms, missing left or right or both legs, and missing one arm and one leg. For each pattern, we create a pair of query/index sets, such that all the query samples have the designated visibility pattern, while all the index samples always have full visibility. We evaluate models on these $10$ sets and report the averaged results.

We compare the two proposed keypoint occlusion augmentation training strategies, i.e., independent keypoint dropout and structured keypoint dropout, along with two other baselines, one (``None'') that uses uniform (all-one) visibility masks and one (``Thresholding'') that uses visibility masks only from thresholding keypoint detection confidence during training. To demonstrate the upper-bound performance for reference, we further train $10$ models with fixed visibility input, each dedicated to one of the $10$ targeted occlusion patterns. Each individual model is evaluated on its dedicated visibility pattern.

\paragraph{Real-Distribution Occlusion} We test model performance with the top-$50$ most frequent visibility patterns from the in-the-wild pose keypoint joint distribution introduced in Section~\ref{sec:keypoint_availability_aug}, which accumulatively covers about $81\%$ of the visibility patterns of the $300$ million poses. For each pattern, we create a pair of query/index sets, such that all the query samples have the designated visibility pattern. For every index sample, we randomly draw a visibility pattern from the joint distribution whose visible keypoints make a superset of the query visible keypoints. We evaluate models on these $50$ sets and report the average results weighted by the frequency of each visibility pattern. It is also worth mentioning that only $5$ of the $10$ targeted occlusion patterns are among the top-$50$ in-the-wild patterns.

We additionally experiment with 3DPW. 3DPW provides realistic occlusions on 2D keypoints, which we directly use for masking our model input. We also smear these occlusions to 3D keypoints and only use visible 3D keypoints for NP-MPJPE computation during evaluation.

Using these datasets, we explore Pr-VIPE performance with different keypoint occlusion augmentation strategies described in Section~\ref{sec:occlusion_robust_embedding}. 

\begin{table*}
  \centering
  \caption{Comparison of cross-view pose retrieval results Hit@$k$ ($\%$) on H3.6M with synthetic occlusions using different keypoint occlusion augmentation training strategies. $^{\dagger}$ indicates the results are from dedicated models that are trained for each visibility pattern and evaluated with query/index samples with identical keypoint visibilities.} \label{tab:partial_h36m}
  \scalebox{1.0}{
   \begin{tabular}{c c | c c c c }
   \toprule[0.2em]

   Evaluation type & Keypoint dropout & $k=1$ & $k=5$ & $k=10$ & $k=20$ \\

   \toprule[0.2em]
  \multirow{4}{*}{No occlusion} & None & $\textbf{72.9}$ & $\textbf{90.0}$ & $\textbf{94.0}$ & $\textbf{96.6}$ \\
  & Thresholding & $72.5$ & $89.9$ & $93.8$ & $96.5$ \\
  & Independent & $71.7$ & $89.7$ & $93.7$ & $96.4$ \\
  & Structured & $71.6$ & $89.5$ & $93.6$ & $96.2$ \\
  \hline
  \multirow{5}{*}{Targeted occlusion} & None & $2.52$ & $5.71$ & $7.51$ & $10.5$ \\
  & Thresholding & $17.0$ & $32.9$ & $41.5$ & $51.2$ \\  
  & Independent & $66.2$ & $\textbf{87.3}$ & $\textbf{92.1}$ & $\textbf{95.3}$ \\
  & Structured & $\textbf{66.5}$ & $\textbf{87.3}$ & $92.0$ & $95.2$ \\\cline{2-6}
  & Dedicated$^{\dagger}$ & $75.7$ & $91.5$ & $94.7$ & $96.8$ \\
  \hline
  \multirow{4}{*}{Real-distribution occlusion}  & None & $46.6$ & $63.5$ & $69.6$ & $75.1$ \\
  & Thresholding & $58.4$ & $80.4$ & $86.6$ & $91.3$ \\  
  & Independent & $70.3$ & $89.2$ & $93.5$ & $\textbf{96.2}$ \\
  & Structured & $\textbf{71.0}$ & $\textbf{89.5}$ & $\textbf{93.6}$ & $\textbf{96.2}$ \\  
   \bottomrule[0.1em]
\end{tabular}
}
\end{table*}

\begin{table}
  \centering
  \caption{Comparison of cross-view pose retrieval results Hit@$k$ ($\%$) on 3DPW with realistic occlusions using different keypoint occlusion augmentation training strategies.} \label{tab:partial_3dpw}
  \scalebox{1.0}{
   \begin{tabular}{c | c c c c }
   \toprule[0.2em]
    Keypoint dropout & $k=1$ & $k=5$ & $k=10$ & $k=20$ \\
   \toprule[0.2em]
  None &  $43.8$ & $62.0$ & $69.5$ & $76.7$ \\
  Thresholding & $62.4$ & $85.3$ & $91.1$ & $95.0$ \\  
  Independent & $\textbf{64.6}$ & $\textbf{87.2}$ & $\textbf{92.9}$ & $\textbf{96.2}$ \\
  Structured & $63.9$ & $86.2$ & $92.6$ & $95.8$\\  
   \bottomrule[0.1em]
\end{tabular}
}
\end{table}

\begin{figure*}
  \centering
  \includegraphics[width=\textwidth]{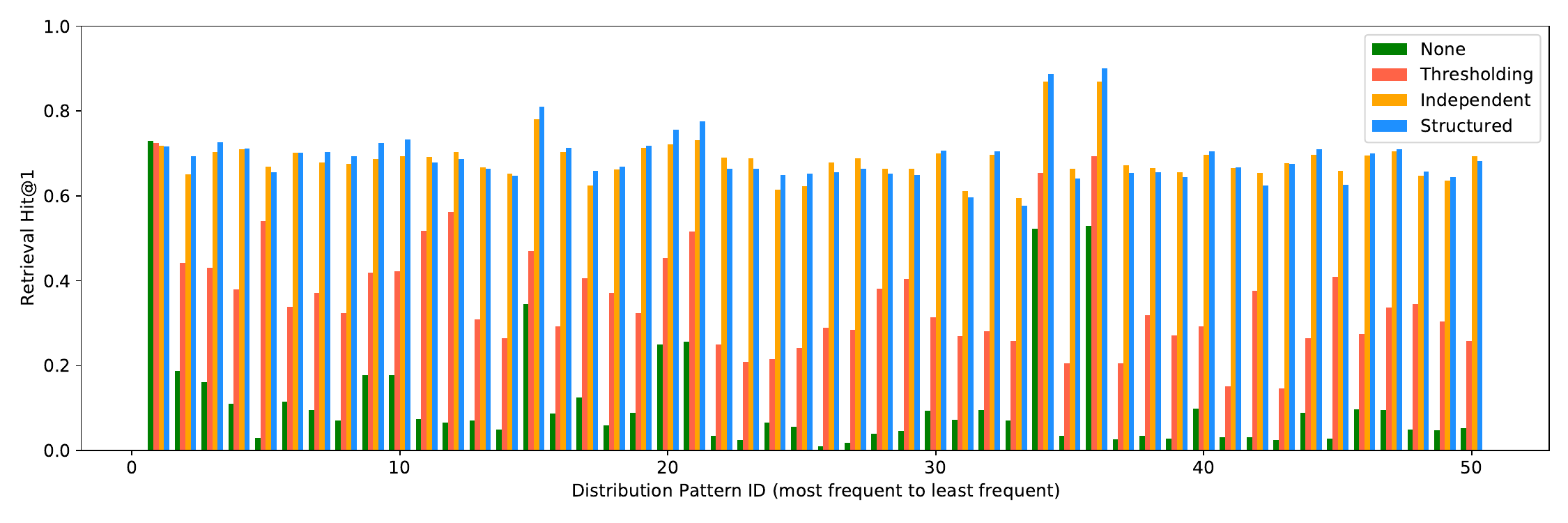}
  \caption{Comparison of cross-view pose retrieval results Hit@$1$ for the top-$50$ most common in-the-wild occlusion patterns using different keypoint dropout methods with Pr-VIPE.}
 \label{fig:barplot_50}
\end{figure*}

\subsubsection{Quantitative Results}

Table~\ref{tab:partial_h36m} shows pose retrieval results on H3.6M synthetic test sets. We see that when training without considering occlusions (``None''), the model performance is significantly lowered when occlusions exist at test time, compared to when no occlusions exist. Relying on limited realistic occlusions from the training set (``Thresholding'') can boost models' robustness to occlusions as compared to without. The proposed independent or the structured keypoint dropout strategy further provides significant performance improvement under different occlusion scenarios, with only slightly lower performance on fully visible retrieval. Similarly in Table~\ref{tab:partial_3dpw}, we see that using our proposed keypoint dropout strategies outperform baseline methods by a large margin. The results on the outdoor 3DPW dataset also suggest that our occlusion-robust Pr-VIPE model works in the wild.

For targeted occlusions, the proposed one-for-all models using independent or structured keypoint dropout achieve strong performance compared with the dedicated model upper bounds on several retrieval metrics. Though there is a performance difference between one-for-all models and dedicated models, it is important to note that each dedicated model is limited only to one specific visibility pattern and not robust to other occlusion patterns. In practice, it is infeasible to train and deploy a model for every possible visibility pattern.

For targeted and real-distribution occlusions, we observe that using structured dropout outperforms independent dropout (Table~\ref{tab:partial_h36m}). From a modeling perspective, structured dropout is more flexible than independent dropout at incorporating prior occlusion pattern statistics into training, and thus has better performance on similarly distributed test time patterns.

Fig.~\ref{fig:barplot_50} further shows the retrieval results for each of the top-$50$ most common in-the-wild visibility patterns. 
We see that on the fully visible pose represented by the first bar, all the dropout methods are similar in performance. 
As we introduce different occlusion patterns, independent and structured dropout performs much better than the other baselines. 
In conclusion, training with keypoint occlusion augmentation is important to improving model robustness to partial visibilities.

\subsubsection{Qualitative Results}

\def\figsize{0.15}
\def\fighspace{-2mm}
\def\fighspacer{+1.5mm}
\begin{figure*}
\centering
\begin{tabular}{cccccc}
\centering
\scriptsize{$C=0.727$}\hspace{\fighspace} & \scriptsize{$E=0.09$}\hspace{\fighspacer} &
\scriptsize{$C=0.621$}\hspace{\fighspace} & \scriptsize{$E=0.08$}\hspace{\fighspacer} & 
\scriptsize{$C=0.866$}\hspace{\fighspace} & \scriptsize{$E=0.08$}\hspace{\fighspacer}\\
\includegraphics[width=\figsize\textwidth]{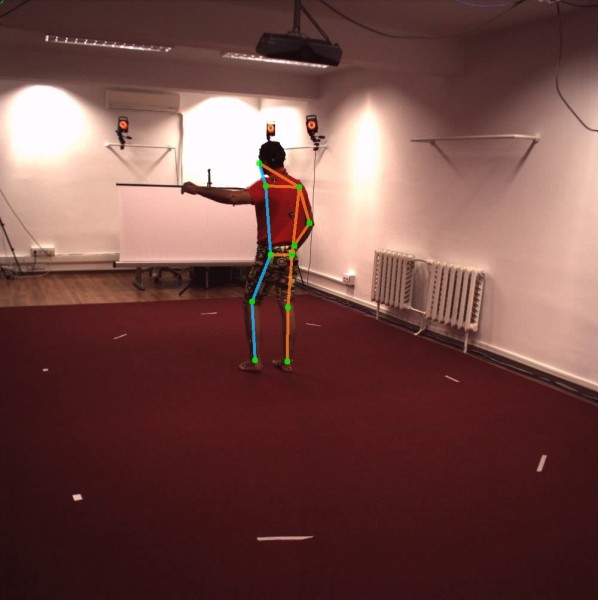}\hspace{\fighspace} & \includegraphics[width=\figsize\textwidth]{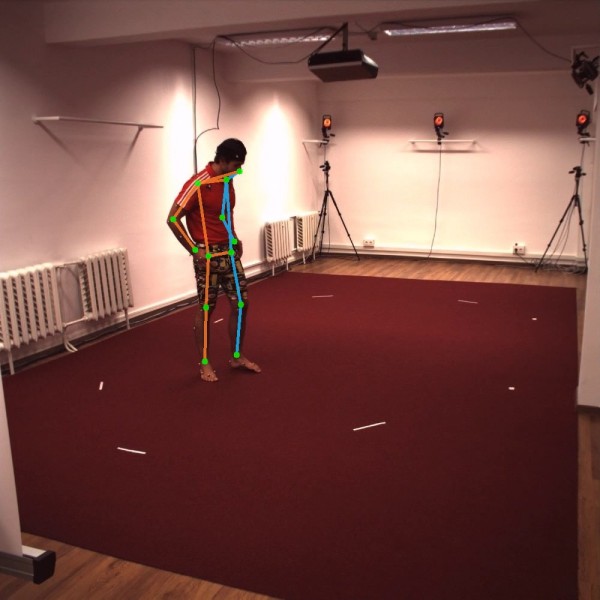}\hspace{\fighspacer} &
\includegraphics[width=\figsize\textwidth]{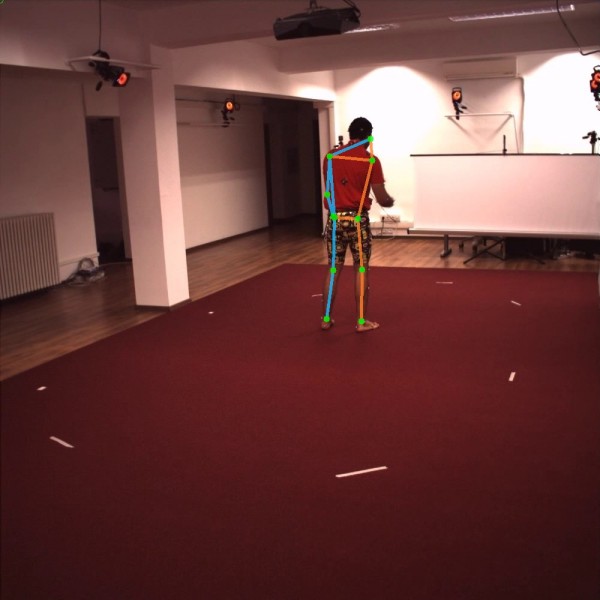}\hspace{\fighspace} & \includegraphics[width=\figsize\textwidth]{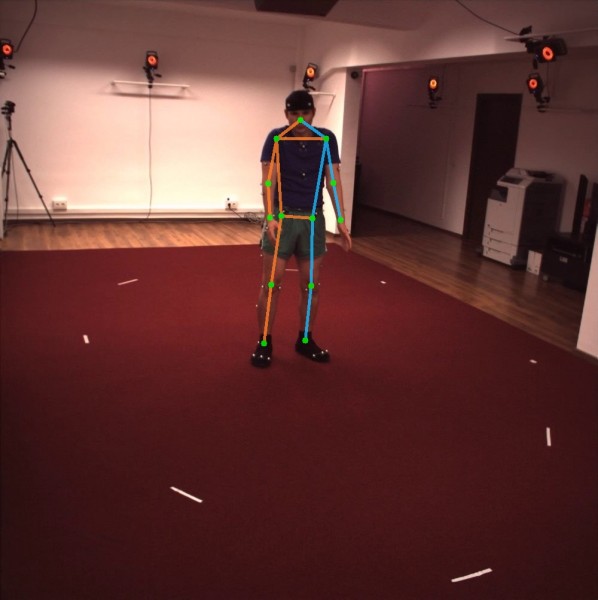}\hspace{\fighspacer}  &
\includegraphics[width=\figsize\textwidth]{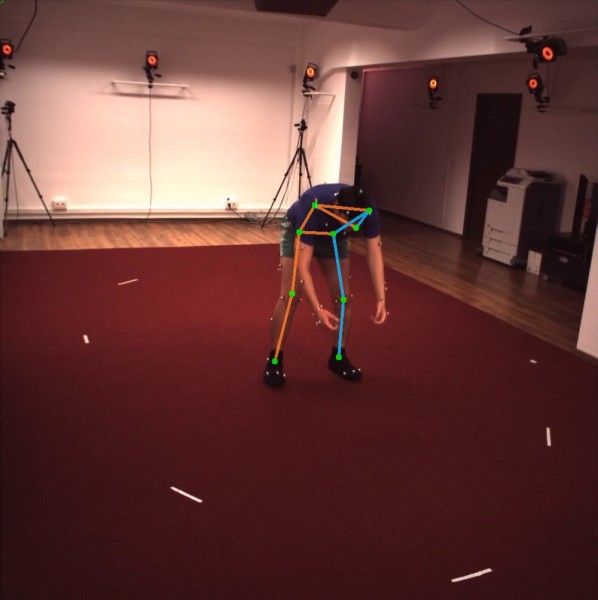}\hspace{\fighspace} & \includegraphics[width=\figsize\textwidth]{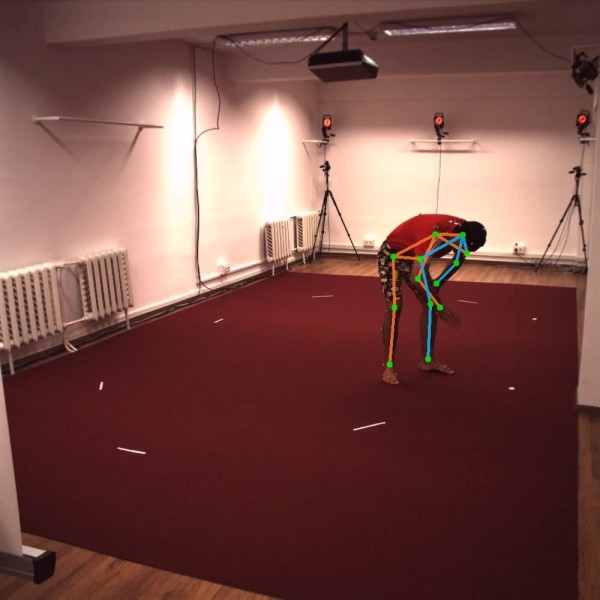}\hspace{\fighspacer}  \\
\scriptsize{(Left arm)} & & \scriptsize{(Right arm)} & & \scriptsize{(Both arms)} & \\
& & & & & \\
\scriptsize{$C=0.943$}\hspace{\fighspace} & \scriptsize{$E=0.06$}\hspace{\fighspacer} &
\scriptsize{$C=0.862$}\hspace{\fighspace} & \scriptsize{$E=0.06$}\hspace{\fighspacer}  &
\scriptsize{$C=0.682$}\hspace{\fighspace} & \scriptsize{$E=0.08$}\hspace{\fighspacer} \\
\includegraphics[width=\figsize\textwidth]{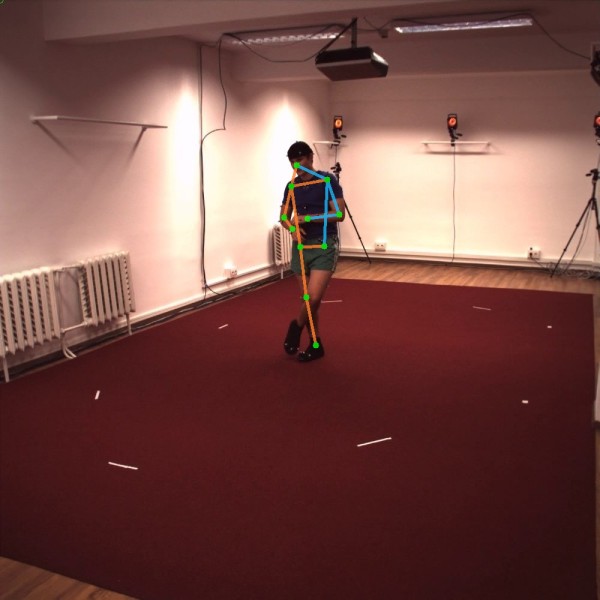}\hspace{\fighspace} & \includegraphics[width=\figsize\textwidth]{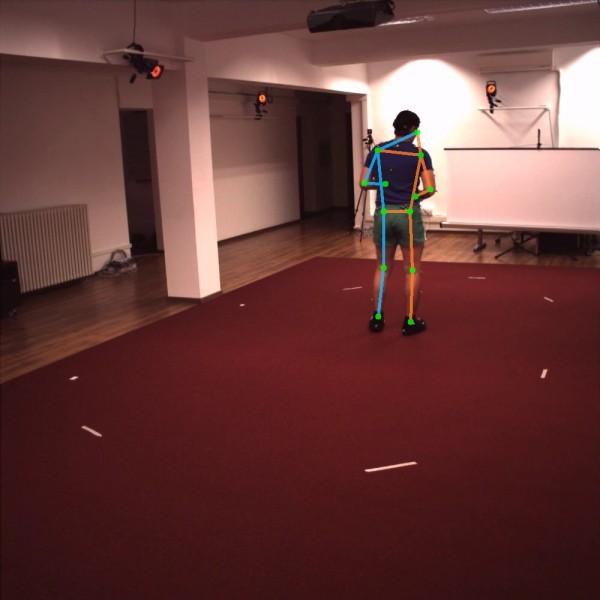}\hspace{\fighspacer} & \includegraphics[width=\figsize\textwidth]{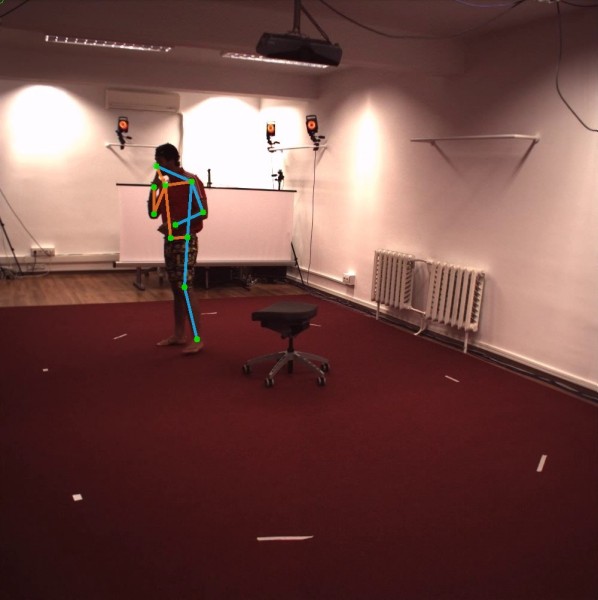}\hspace{\fighspace} & \includegraphics[width=\figsize\textwidth]{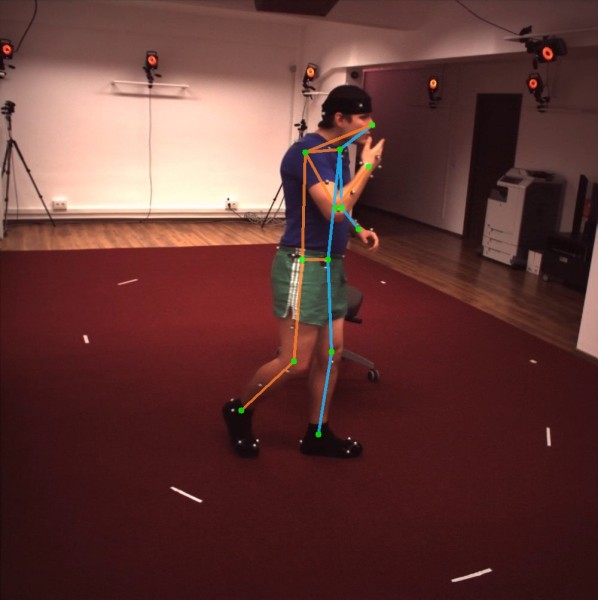}\hspace{\fighspacer} & \includegraphics[width=\figsize\textwidth]{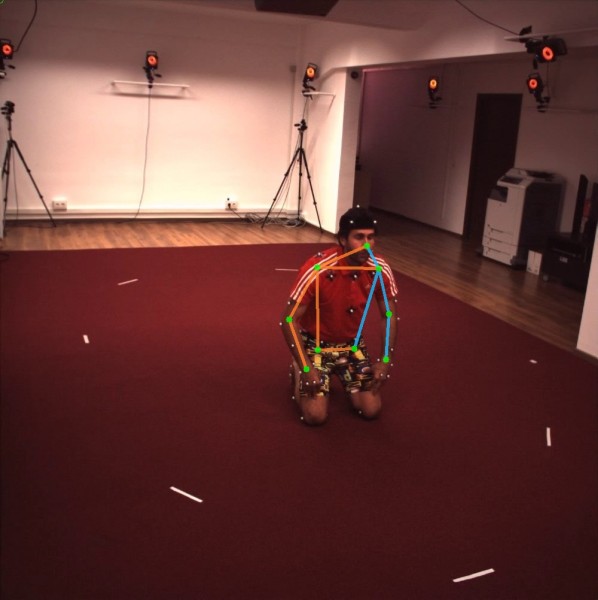}\hspace{\fighspace} & \includegraphics[width=\figsize\textwidth]{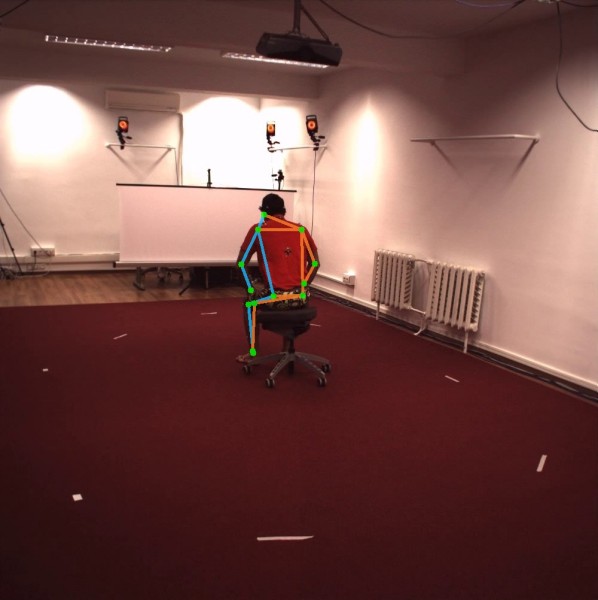}\hspace{\fighspacer}  \\
\scriptsize{(Left leg)} & & \scriptsize{(Right leg)} & & \scriptsize{(Both legs)} & \\

\end{tabular}
\caption{Visualization of keypoint subset pose retrieval. For each pose pair, we show the query pose on the left and the top-$1$ retrieval using the Pr-VIPE model (with camera augmentation and structured keypoint dropout) on the right. The occluded body parts, synthesized with targeted occlusion patterns, are described under each query images. Only visible keypoints are visualized with the skeletons. Retrieval confidences (``$C$'') and top-$1$ 3D NP-MPJPEs (``$E$'') are also displayed for each retrieval.}
\label{fig:partial_retrieval}
\end{figure*}

Fig.~\ref{fig:partial_retrieval} shows qualitative retrieval results on 6 targeted occlusion patterns using Pr-VIPE (with camera augmentation) trained with structured keypoint dropout. The queries are synthetically occluded. Specifically, queries have left, right, and both arms occluded, respectively, in the top row, and left, right, and both legs occluded, respectively, in the bottom row.
Using our embeddings, we are able to accurately retrieve samples that share similar poses in terms of the visible keypoints.
For example, in the bottom right pair (with query having both legs occluded), the top-$1$ retrieval has the same upper body pose as the query, despite its significant difference in the lower body pose.
This evaluation also indicates that using our occlusion-robust embedding, one will be able to specify a subset of query keypoints for targeted partial pose retrieval.

\subsection{Action Recognition}\label{sec:action_recognition}

Our pose embedding can be directly applied to pose-based downstream tasks using simple algorithms. Here, we compare the performance of Pr-VIPE (\textbf{only trained on H3.6M, with no additional training}) on the Penn Action dataset against other approaches specifically trained for action recognition on the target dataset.

\subsubsection{Evaluation Procedure}\label{sec:action_recognition_procedure}

To apply our frame embedding to videos, we compute Pr-VIPEs on single video frames and use the negative logarithm of the matching probability (\ref{eq:9}) as the distance between two frames. Using this per frame distance, we apply temporal averaging within an atrous kernel of size $7$ and rate $3$ around the two center frames and use this averaged distance as the frame matching distance in a sequence. Given the matching distance, we use the standard dynamic time warping (DTW) algorithm to align two pose sequences by minimizing the sum of frame matching distances. The averaged frame matching distance from the DTW alignment of Pr-VIPE matching distance is used as the distance between two video sequences.

For Temporal Pr-VIPE, our procedure is similar to above, except we do not stack embeddings, as the model directly embeds sequences. We compute sequence embeddings centered at each frame over the same temporal windows as above, and use DTW over sequence embedding distances to align video pairs. The distance between two video sequences is computed based on DTW-aligned sequences the same as above.

\begin{table}[t]
    \centering
    \caption{Comparison of action recognition results on Penn Action.}\label{tab:action_recognition}
    \scalebox{0.95}{
    \begin{tabular}{l|ccc|c}
        \toprule[0.2em]
        \multirow{2}{*}{Methods} & \multicolumn{3}{c|}{Input} & Accuracy\\
        & RGB & Flow & Pose & ($\%$) \\
        \toprule[0.2em]
        Nie~\textit{et al.}~\cite{nie2015joint} & \checkmark & & \checkmark & $85.5$ \\
        Iqbal~\textit{et al.}~\cite{iqbal2017pose} & & & \checkmark & $79.0$\\
        Cao~\textit{et al.}~\cite{cao2017body} & & \checkmark & \checkmark & $95.3$ \\
        & \checkmark & \checkmark & & $98.1$ \\
        Du~\textit{et al.}~\cite{du2017rpan} & \checkmark & \checkmark & \checkmark & $97.4$ \\
        Liu~\textit{et al.}~\cite{liu2018recognizing} & \checkmark & & \checkmark & $91.4$ \\
        Luvizon~\textit{et al.}~\cite{luvizon2019multi} & \checkmark & & \checkmark & $98.7$ \\
        \bottomrule[0.1em]
        \textbf{Ours:} & & & &\\
        Pr-VIPE (stacking 8D) & & & \checkmark & $95.3$\\        
        Pr-VIPE (stacking 16D) & & & \checkmark & $97.4$\\
        Pr-VIPE (stacking 32D) & & & \checkmark & $98.0$\\        
        Temporal Pr-VIPE (32D) & & & \checkmark & $98.0$\\
        Temporal Pr-VIPE (56D) & & & \checkmark & $97.8$\\        
        \bottomrule[0.1em]
        \textbf{Ours (1-view index):} & & & &\\
        Pr-VIPE (stacking 8D) & & & \checkmark & $90.0$\\
        Pr-VIPE (stacking 16D) & & & \checkmark & $91.2$\\
        Pr-VIPE (stacking 32D) & & & \checkmark & $92.7$\\        
        Temporal Pr-VIPE (32D) & & & \checkmark & $93.3$\\
        Temporal Pr-VIPE (56D) & & & \checkmark & $93.7$\\        
        \bottomrule[0.1em]
    \end{tabular}
    }
\end{table}

\begin{table}[t]
    \centering
    \caption{Comparison of video alignment results on Penn Action.} \label{tab:seq_align_res}
      \begin{tabular}{l | c  }
   \toprule[0.2em]
  Methods &  Kendall's Tau \\
   \toprule[0.2em]
   SaL \cite{misra2016shuffle} & $0.6336$\\
   TCN \cite{sermanet2018time} & $0.7353$\\
   TCC \cite{dwibedi2019temporal} & $0.7328$\\
   TCC + SaL \cite{dwibedi2019temporal} & $0.7286$\\
   TCC + TCN \cite{dwibedi2019temporal} & $0.7672$\\
   \bottomrule[0.1em] 
    \textbf{Ours:} & \\
    Pr-VIPE (stacking 8D) &  $0.7071$\\
    Pr-VIPE (stacking 16D) &  $0.7319$\\
    Pr-VIPE (stacking 32D) &  $0.7405$\\    
    Temporal Pr-VIPE (32D) & $0.7630$\\
    Temporal Pr-VIPE (56D) & $0.7810$\\    
    \bottomrule[0.1em] 
    \textbf{Ours (same-view only):} & \\
    Pr-VIPE (stacking 8D) &  $0.7077$\\
    Pr-VIPE (stacking 16D) &  $0.7331$\\
    Pr-VIPE (stacking 32D) &  $0.7419$\\    
    Temporal Pr-VIPE (32D) & $0.7639$\\ 
    Temporal Pr-VIPE (56D) & $0.7847$\\     
    \bottomrule[0.1em] 
    \textbf{Ours (different-view only):} & \\
    Pr-VIPE (stacking 8D) &  $0.7186$\\
    Pr-VIPE (stacking 16D) &  $0.7346$\\
    Pr-VIPE (stacking 32D) &  $0.7461$\\    
    Temporal Pr-VIPE (32D) & $0.7763$\\
    Temporal Pr-VIPE (56D) & $0.7934$\\    
    \bottomrule[0.1em]
\end{tabular}
\end{table}

\begin{figure*}
  \centering
  \includegraphics[width=\textwidth]{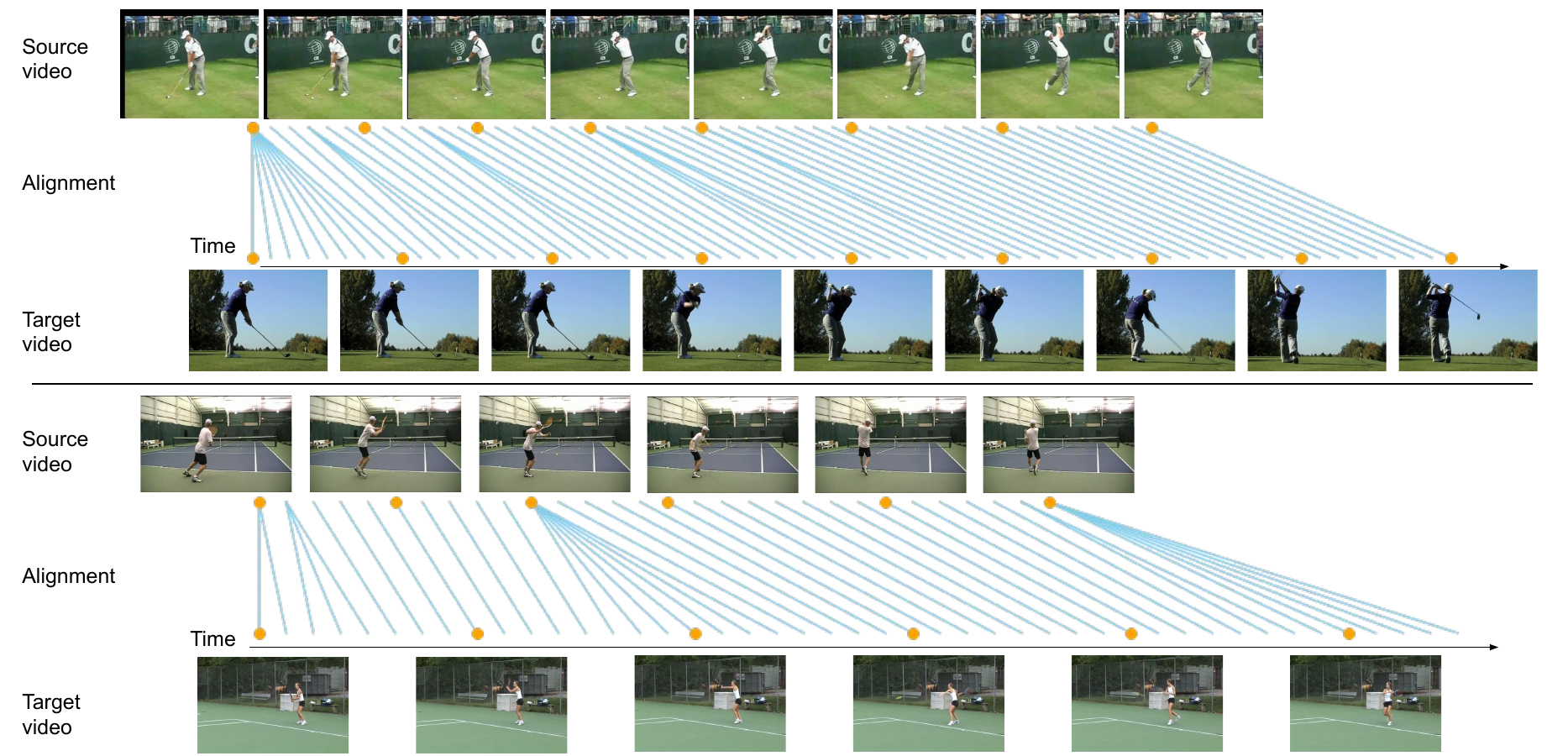}
  \caption{Video alignment results using Pr-VIPE. The orange dots correspond to the visualized frames, and the blue line segments illustrate the frame alignment. }
 \label{fig:video_sync}
\end{figure*}

We evaluate our embeddings for action recognition using nearest neighbor search with the sequence distance described above. Using all the testing videos as queries, we conduct two experiments: (1) we use all training videos as index to evaluate overall performance and compare with state-of-the-art methods, and (2) we use training videos only under one view as index and evaluate the effectiveness of our embeddings in terms of view invariance. For this second experiment, actions with zero or only one sample under the index view are ignored, and accuracy is averaged over different views.

We follow the standard evaluation protocol~\cite{nie2015joint} and remove action ``strum guitar'' and several videos in which less than one third of the target person is visible. We use the official train/test split and report the averaged per-class accuracy. For the view-invariant action recognition experiments in which the index set only contains videos from a single view, we exclude the actions that have zero or only one sample under a particular view. We take the bounding boxes provided with the dataset and use~\cite{papandreou2017towards} (ResNet101) for 2D pose keypoint estimation. For frames of which the bounding box is missing, we copy the bounding box from a nearest frame. Finally, since our embedding is chiral, but certain actions can be done with either body side (pitching a baseball with left or right hand), when we compare two frames, we extract our embeddings from both the original and the mirrored version of each frame, and use the minimum distance between all the pairwise combinations as the frame distance.

\subsubsection{Results}

Table~\ref{tab:action_recognition} demonstrates that our embeddings can achieve competitive results on pose-based action classification without any training of the Pr-VIPE model on the target domain or using image context information. Pr-VIPE with nearest neighbor retrieval is able to outperform the existing best baseline that only uses pose input, as well as other methods that has access to input image context or optical flow. This result shows that our embedding is applicable to pose-based action recognition without additional training, using the method described in Section~\ref{sec:action_recognition_procedure} for pose sequences. 

In particular, we note that Temporal Pr-VIPE is able to classify actions more accurately as compared to stacking frame-level Pr-VIPE (Table~\ref{tab:action_recognition}). Stacking frame-level embeddings also increases dimensionality of the embeddings (for example, $7 \times 16 = 112$D), whereas temporal embeddings achieves higher classification accuracy using only 32 dimensions.  

To evaluate the view-invariance property of our embeddings, we perform action classification using only training videos from a single view. The last section in Table~\ref{tab:action_recognition} further demonstrates the advantages of our view-invariant embeddings, as they can be used to classify actions from different views using index samples from only one single view with relatively high accuracy.

\subsection{Video Alignment}\label{sec:sequence_alignment}

The view-invariance property of Pr-VIPE can be leveraged to align videos with pose sequences across views. We apply Pr-VIPE (\textbf{only trained on H3.6M, with no additional training}) on the Penn Action dataset against other approaches specifically trained for video alignment on the target dataset.

\subsubsection{Evaluation Procedure}

Our embeddings can be used to align human action videos from different views using DTW algorithm. Similar to the temporal embedding stacking described in Section~\ref{sec:action_recognition_procedure}, we apply temporal averaging of the negative logarithm of matching probability within an atrous kernel of size $7$ and rate $3$ around each frame in a pair of videos. This averaged distance is used as the frame matching distance in a sequence and minimized using DTW to align two pose sequences. 
For temporal Pr-VIPE, we compute sequence embeddings on the same frame windows, and the frame matching distance is then computed between these sequence embeddings.

We follow~\cite{dwibedi2019temporal} and measure the alignment quality of our embeddings quantitatively using Kendall's Tau~\cite{kendall1938new}, which reflects how well an embedding model can be applied to align unseen sequences if we use nearest neighbor in the embedding space to match frames for video pairs. A value of $1$ corresponds to perfect alignment. We also test the view-invariant properties of our embeddings by evaluating Kendall's Tau on aligning videos pairs from the same view, and aligning pairs with different views.

We follow the protocol in~\cite{dwibedi2019temporal}, excluding ``jump rope'' and ``strum guitar'' from our evaluation. For the evaluations between videos under only the same or different views, we exclude actions that have zero videos under a particular view from the average Kendall's Tau computation. Since certain actions can be done with either body side, for a video pair $(v_1,v_2)$, we compute the Kendall's Taus between $(v_1,v_2)$ and $(v_1,\text{mirror}(v_2))$, and take the larger value.

\subsubsection{Results}

In Table~\ref{tab:seq_align_res}, we compare our results with other video embedding baselines that are trained for the alignment task on Penn Action. Results show that all Pr-VIPE methods outperform SaL, while Pr-VIPE (stacking 16D and stacking 32D) performs comparable to all the method that use a single type of loss. While Pr-VIPE tau is lower than the combined TCC+TCN loss, our embeddings are able to achieve this without being explicitly trained for this task or taking advantage of image context. This demonstrates that Pr-VIPE is applicable to aligning videos containing pose sequences. 

Our experiments with Temporal Pr-VIPE show that it performs better than stacking Pr-VIPE at alignment. Additionally, $56$D temporal Pr-VIPE is able to outperform all alignment baselines. We note that temporal embeddings are able to achieve this with much lower embedding dimension compared to stacking Pr-VIPE, in particular $56$D vs. stacking $32$D (total $224$D). 

In the last two sections of Table~\ref{tab:seq_align_res}, we show the results from aligning video pairs only from the same or different views. We can see that our embeddings achieve consistent performance regardless of whether the aligned video pairs are from the same or different views, which demonstrate their view-invariance property. 
In Fig.~\ref{fig:video_sync}, we show visualization of synchronized action video results from different views using stacking $32$D Pr-VIPE. More aligned videos are available at {\small \url{https://drive.google.com/drive/folders/1nhPuEcX4Lhe6iK3nv84cvSCov2eJ52Xy?usp=sharing}}.

\begin{figure}
  \centering
  \includegraphics[width=0.9\columnwidth]{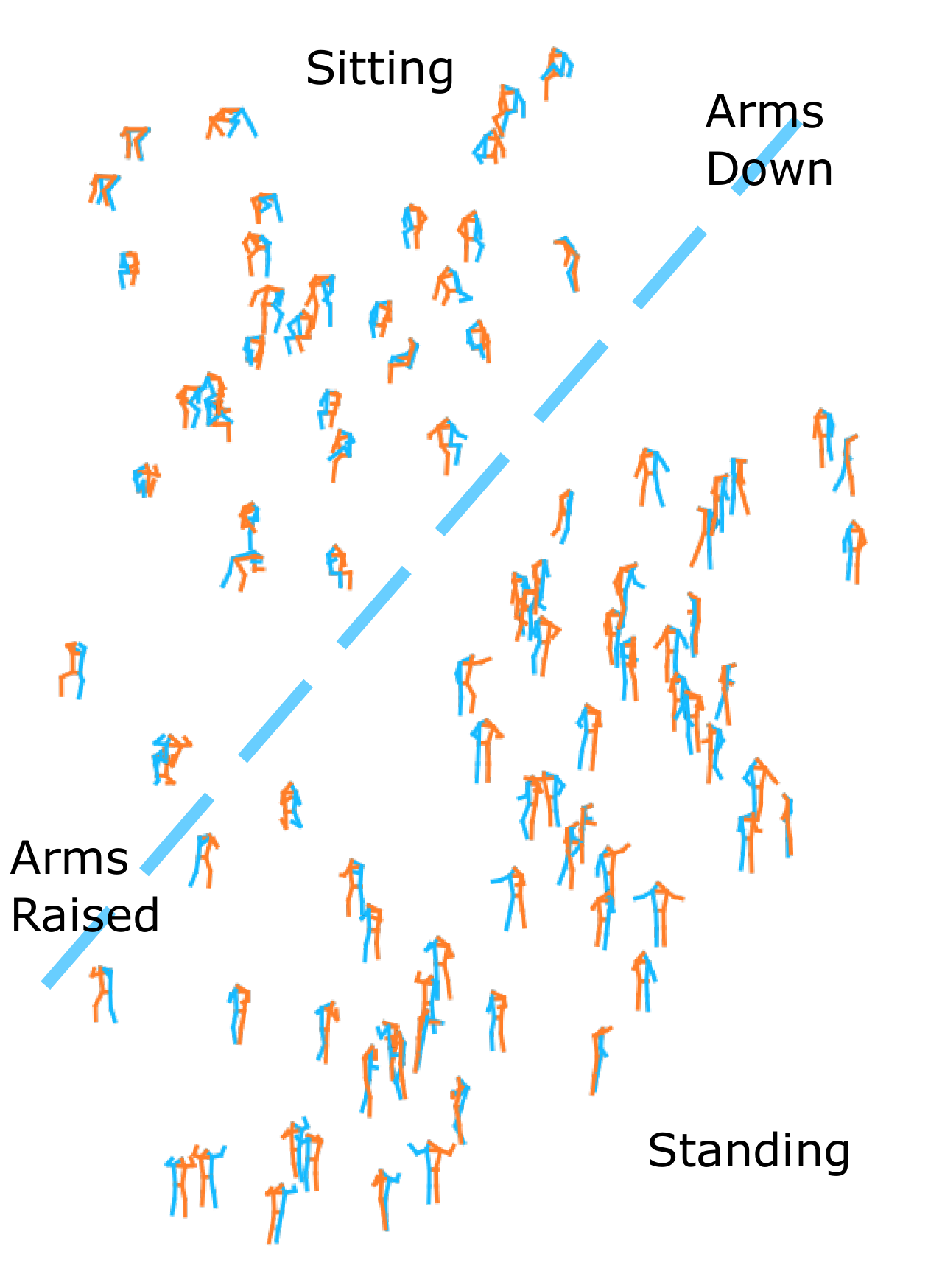}
  \caption{Visualization of Pr-VIPE space with 2D poses in the H3.6M hold-out subset using the first two PCA dimensions. The dotted line shows the separation between sitting and standing poses.}
 \label{fig:pca}
\end{figure}


\input{5.7.embedding_properties}

\input{5.8.ablation_studies}

%% file: 5.7.embedding_properties.tex
\subsection{Embedding Properties}\label{sec:embedding_properties}

\subsubsection{Embedding Space Visualization}

We visualize the Pr-VIPE space using Principal Component Analysis (PCA). The first two principal dimensions of the $16$D Pr-VIPE is shown in Fig.~\ref{fig:pca}. In order to visualize unique poses, we randomly subsample the H3.6M hold-out set and select 3D poses at least $0.1$ NP-MPJPE apart. Fig.~\ref{fig:pca} demonstrates that 2D poses from matching 3D poses are close together, while non-matching poses are farther apart. Standing and sitting poses appear to be well separated from the two principal dimensions. Additionally, there are leaning poses between sitting and standing. Poses near the bottom-left corner of the figure have arms raised, and there is generally a gradual transition to the top-right corner of the figure, where arms are lowered. These results show that from 2D joint keypoints only, we are able to learn view-invariance properties with compact embeddings.

\subsubsection{Effect of Variance}

\begin{figure}
  \centering
  \includegraphics[width=\columnwidth]{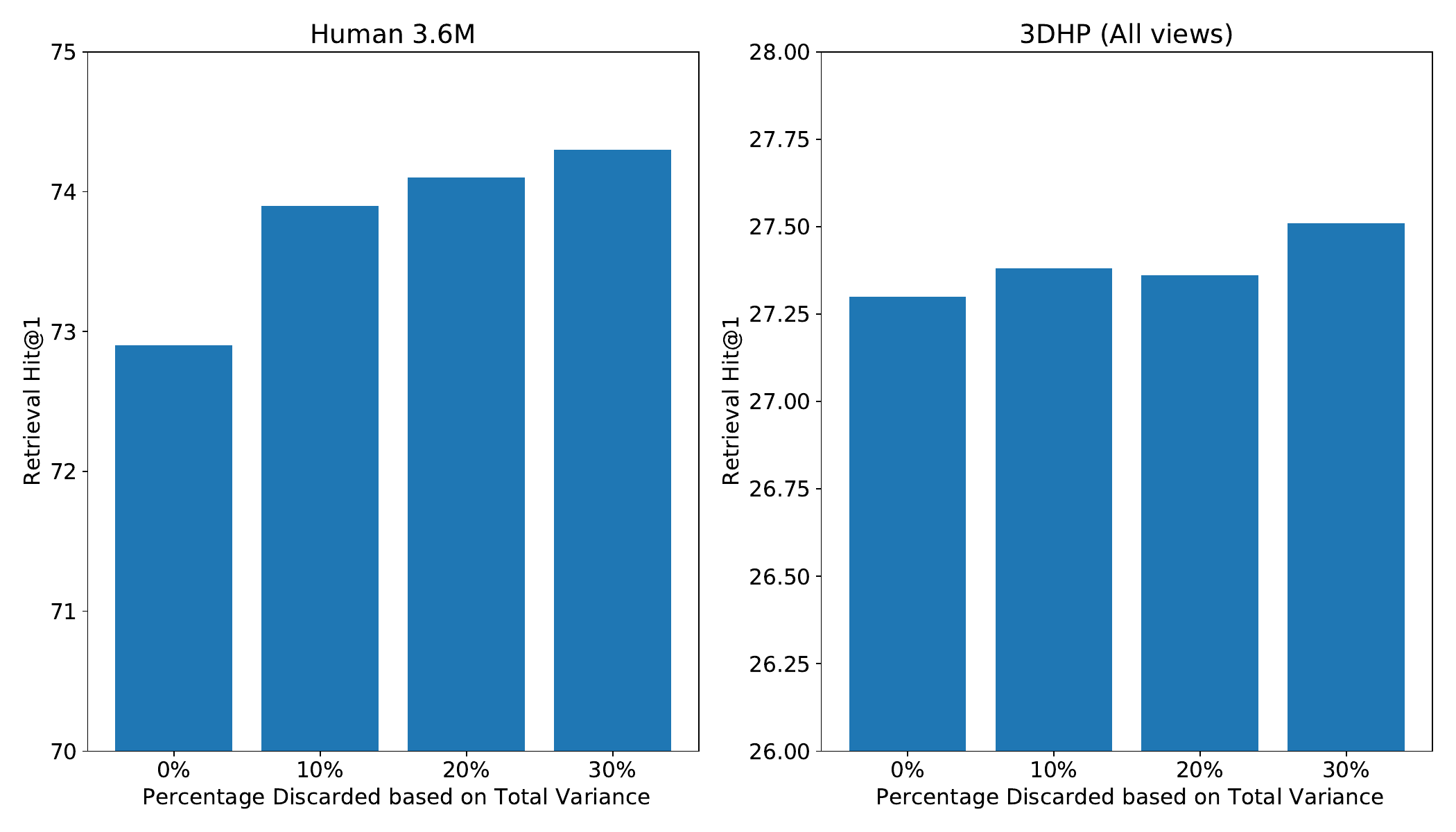}
    \caption{Retrieval Hit@$1$ ($\%$) on H3.6M and 3DHP after discarding samples with largest total variances in query sets. The x-axis denotes the percentage of top sample total variances in query sets.}
 \label{fig:variance_retrieval} 
\end{figure}

We first test the correlation between learned variances and retrieval performance. We rank query poses by their predicted total variance, and discard those with the largest variances. We generally observe increases in retrieval accuracy as we filter out more poses based on variance, as shown in Fig.~\ref{fig:variance_retrieval}.

\begin{figure}
  \centering
  \includegraphics[width=0.9\columnwidth]{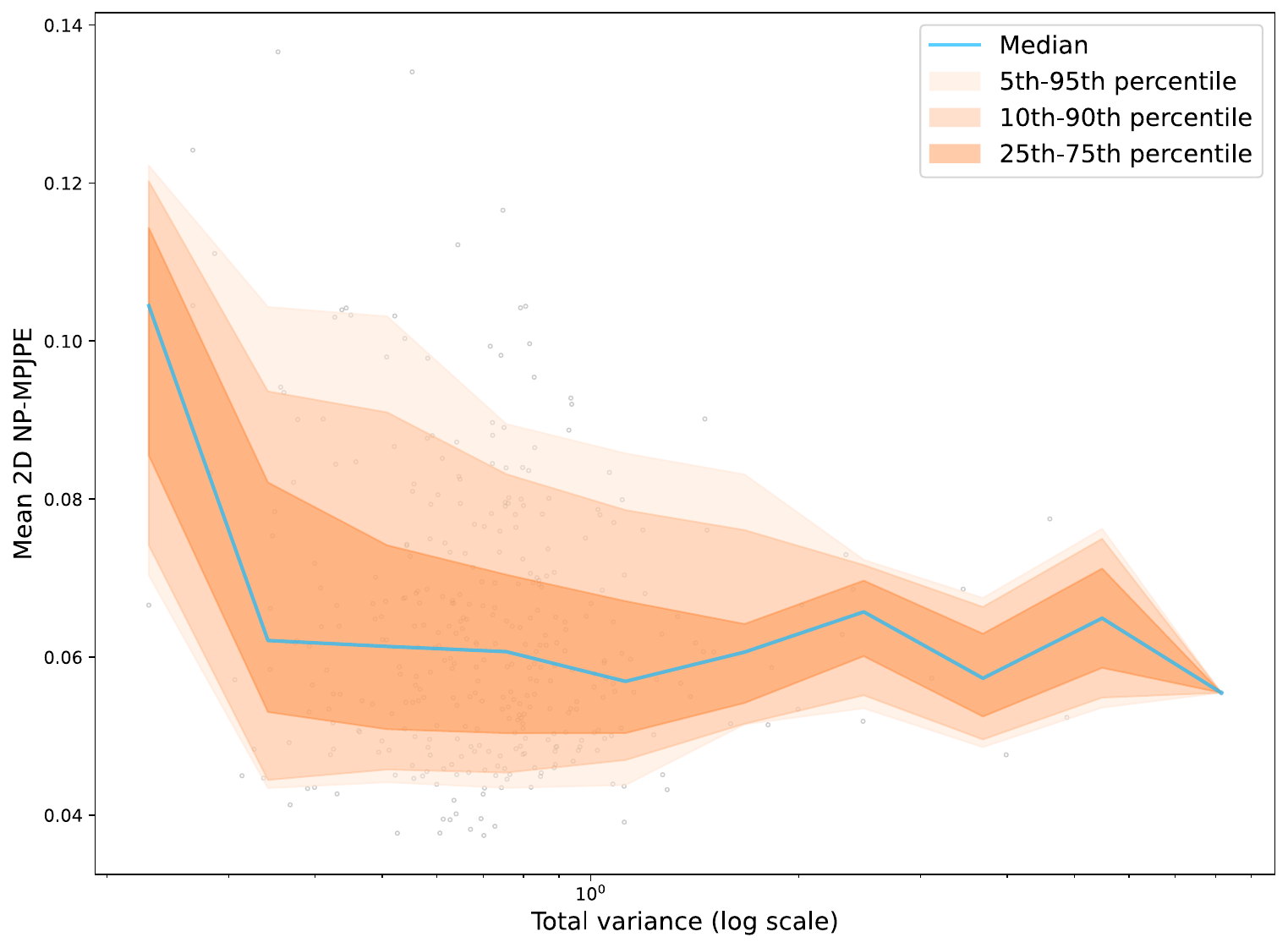}
  \caption{Relationship between embedding variance and averaged 2D NP-MPJPE to top-$10$ nearest 2D pose neighbors from the H3.6M hold-out subset. The median value is plotted as a line with the shaded regions representing different percentile values. Best viewed in color.}
 \label{fig:variance_correlation}
\end{figure}

\begin{figure*}
  \centering
  \subfloat[Poses with top-$5$ largest variance and their nearest neighbors in terms of 2D NP-MPJPE.\label{fig:largest_variance}]{\includegraphics[width=0.9\textwidth]{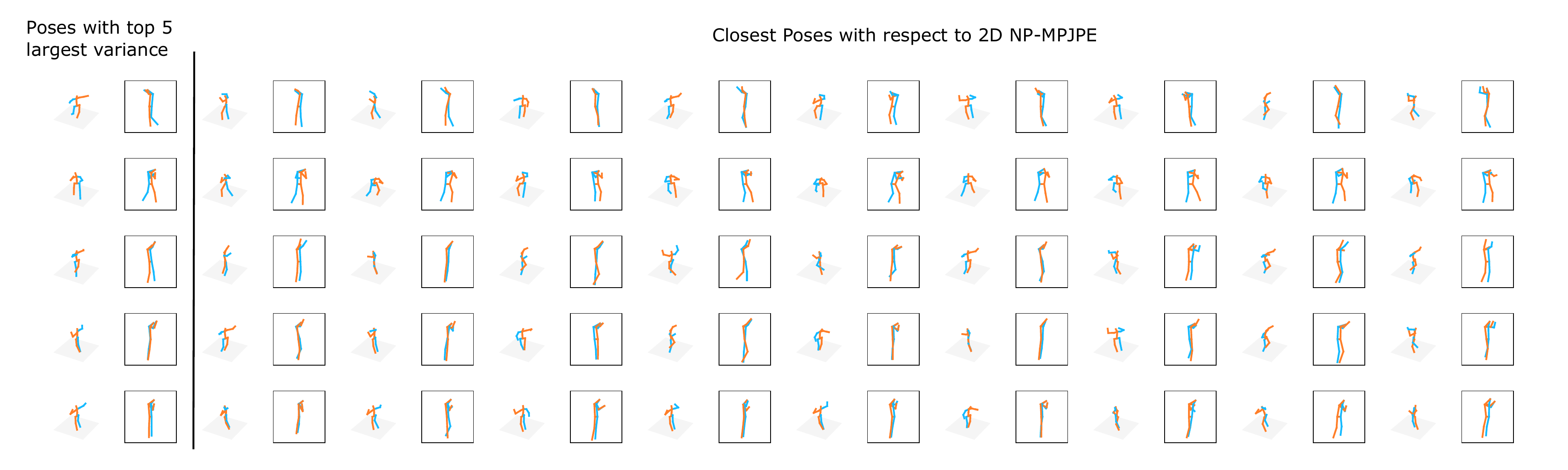}} \\
  \subfloat[Poses with top-$5$ smallest variance and their nearest neighbors in terms of 2D NP-MPJPE.\label{fig:smallest_variance}]{\includegraphics[width=0.9\textwidth]{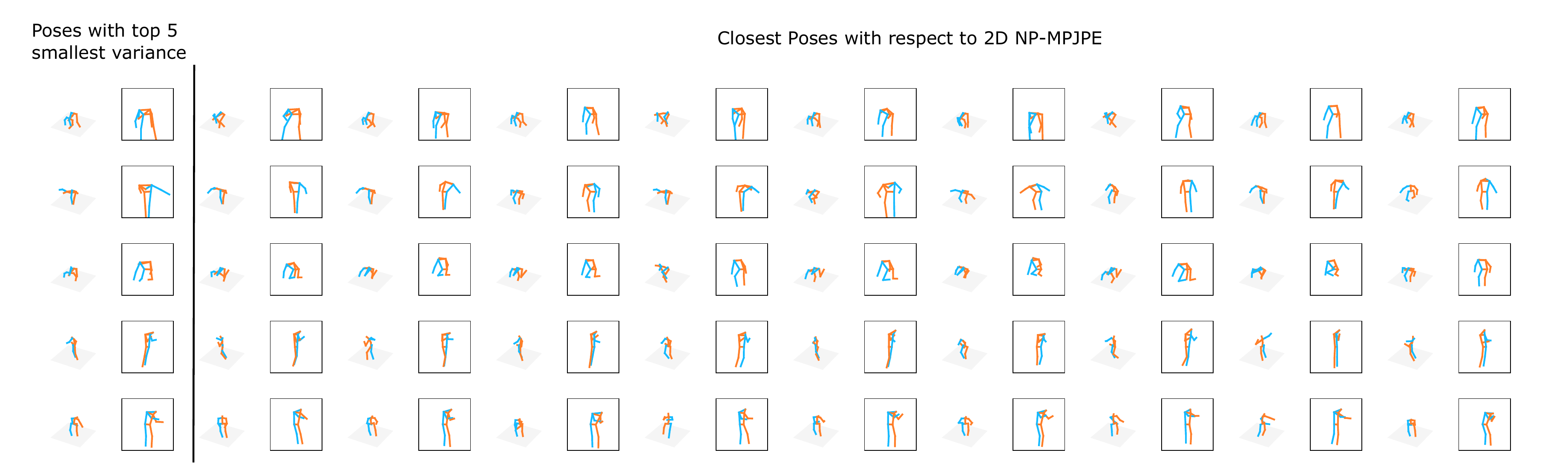}}
\caption{Top retrievals by 2D NP-MPJPE from H3.6M hold-out subset for queries with top-$5$ largest and smallest variances. For each pose pair, the 3D pose is on the left and 2D poses are shown on the right in the boxes.}
\label{fig:variance}
\end{figure*}

Next, we study the relationship between learned variance and input ambiguity in Fig.~\ref{fig:variance_correlation}. This ambiguity is difficult to measure directly, and here we use a heuristic, defined with respect to a dataset (the H3.6M hold-out set). We compute the average 2D NP-MPJPE between a 2D pose and its top-$10$ nearest neighbors in terms of 2D NP-MPJPE. We ensure the 3D poses are different by requiring poses from all camera views to have a minimum gap of $0.1$ 3D NP-MPJPE. If a 2D pose has smaller 2D NP-MPJPE to its closest neighbors, it has more similar 2D poses corresponding to different 3D poses, and thus this 2D pose is more ambiguous. We note that this metric is a heuristic to approximate dataset-dependent 2D ambiguity. To declutter the points in Fig.~\ref{fig:variance_correlation} for clarity, we further deduplicate them by requiring any two samples have 3D NP-MPJPE no less than $0.15$. We observe that generally, there is a decrease in the average 2D NP-MPJPE as the learned variance increases when its value is small, which indicates a negative correlation between the input ambiguity and the variance. We note that this correlation may not be exact, and the embedding uncertainty also depends on other factors, such as model capacity and training data.

Additionally, we sort all the 2D poses used in the above paragraph by their total variance, and Fig.~\ref{fig:largest_variance} shows the 2D poses with the largest variances have more similar 2D poses projected from different 3D poses. In contrast, we see the top-retrieved 2D poses that correspond to the smallest variance poses in Fig.~\ref{fig:smallest_variance} are generally more different.

\subsubsection{Retrieval Confidence}

\begin{figure}
  \centering
  \includegraphics[width=\columnwidth]{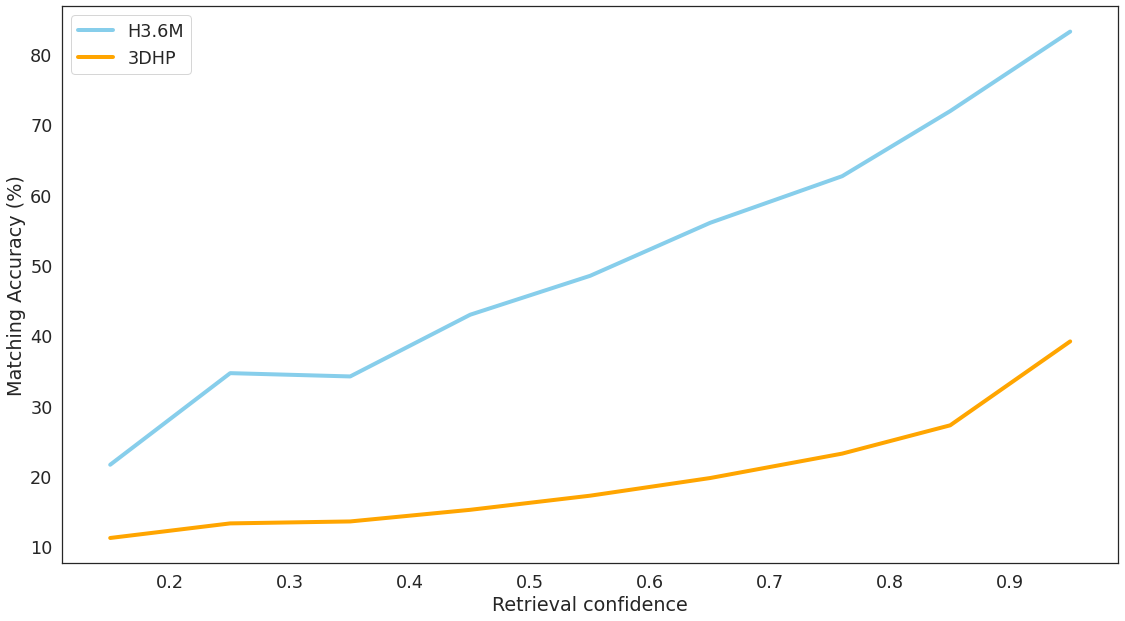}
  \caption{Relationship between retrieval confidence and accuracy using Pr-VIPE (with camera augmentation) on H3.6M and 3DHP.}
 \label{fig:confidence}
\end{figure}

To study the correlation between retrieval confidence and accuracy, we take all the queries along with their top-$5$ retrievals (using Pr-VIPE retrieval confidence) from each query-index camera pair from H3.6M and 3DHP, respectively, and bin each sample by their retrieval confidence. Then we compute average retrieval accuracy in each bin. Fig.~\ref{fig:confidence} shows the matching accuracy for each confidence bin. We can see that the accuracy positively correlates with the confidence values, which suggest our retrieval confidence is a valid indicator to model performance.

%% file: 5.8.ablation_studies.tex
\subsection{Ablation Study}\label{sec:ablation}
\subsubsection{Embedding Dimensions}\label{sec:ablation_dimension}

\begin{figure*}
  \centering
  \includegraphics[width=0.9\textwidth]{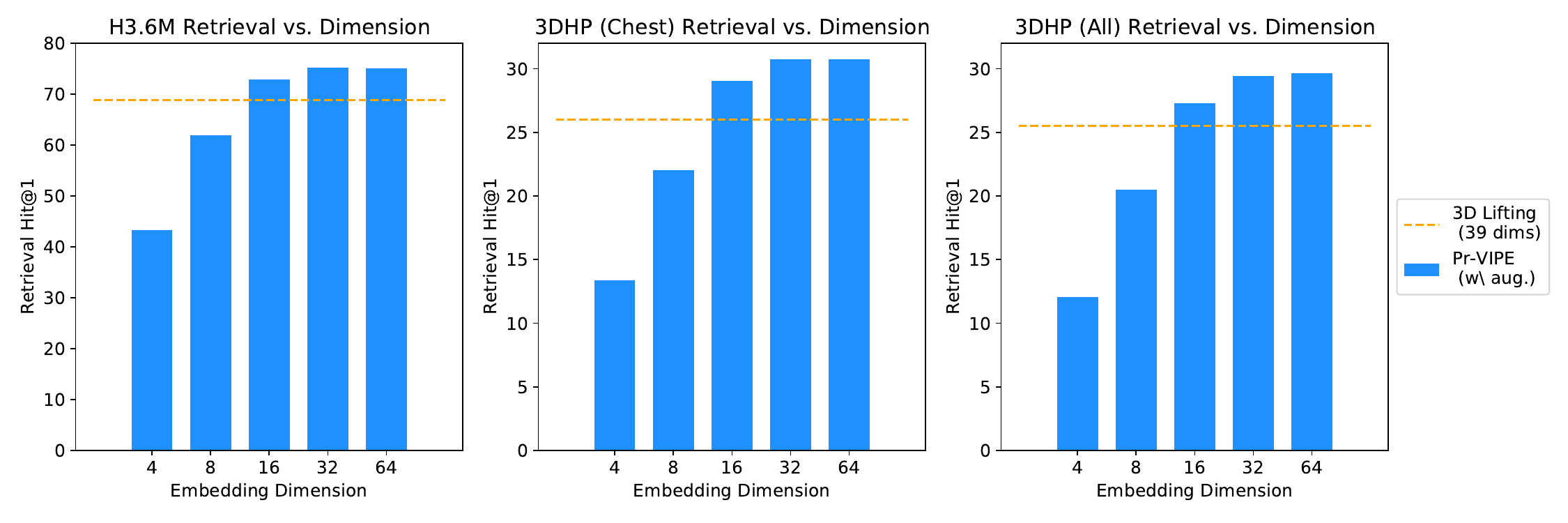}
  \caption{Comparison of retrieval Hit@$1$ with different embedding dimensions on H3.6M and 3DHP. The 3D lifting baseline has output dimension $39$.}
 \label{fig:dimension}
\end{figure*}

Fig.~\ref{fig:dimension} demonstrates the effect of embedding dimensions on retrieval accuracy on H3.6M and 3DHP. The lifting model lifts $13$ 2D keypoints to 3D, and therefore has a constant output dimension of $39$. We see that Pr-VIPE (with camera augmentation) is able to achieve a higher accuracy than lifting at $16$D on all datasets. There is a further increase in accuracy as embedding dimensions increase to $32$D. For all our other experiments, we use dimension $16$.

\subsubsection{What if 2D keypoint detectors were perfect?}

\begin{table}
  \centering
  \caption{Comparison of cross-view pose retrieval results Hit@$1$ ($\%$) using 3D lifting and Pr-VIPE with detected and groundtruth (GT) 2D keypoints on H3.6M and 3DHP.} \label{tab:groundtruth}
  \scalebox{0.9}{
   \begin{tabular}{c c | c c c c c}
   \toprule[0.2em]
    Model & Input type & H3.6M & 3DHP (chest) & 3DHP (all) \\ [0.25ex]
   \toprule[0.2em]
  \multirow{2}{*}{3D lifting} & Detected & $68.8$ & $26.0$ & $25.5$ \\
  & GT & $90.6$ & $52.9$ & $51.0$ \\
   \hline
  \multirow{2}{*}{Pr-VIPE } & Detected & $74.6$ & $25.7$ & $19.8$ \\
  & GT & $97.2$ & $66.6$ & $43.9$ \\   
   \bottomrule[0.1em]
\end{tabular}
}
\end{table}

We repeat our pose retrieval experiments using ground-truth 2D keypoints to simulate a perfect 2D keypoint detector on H3.6M and 3DHP. All experiments use the $4$ views from H3.6M for training following the evaluation procedure in Section~\ref{sec:retrieval_subsection}. Table~\ref{tab:groundtruth} shows the results for the baseline lifting model and Pr-VIPE. These results follow the same trend as using detected keypoints inputs in Table~\ref{tab:3dhp}. Comparing the results of detected and groundtruth keypoints, the large improvement in performance using groundtruth keypoints suggests that a considerable fraction of error from our model is due to imperfect 2D keypoint detections.

\subsubsection{Retrieval across different camera pairs}

\begin{figure}
  \centering
  \includegraphics[width=\columnwidth]{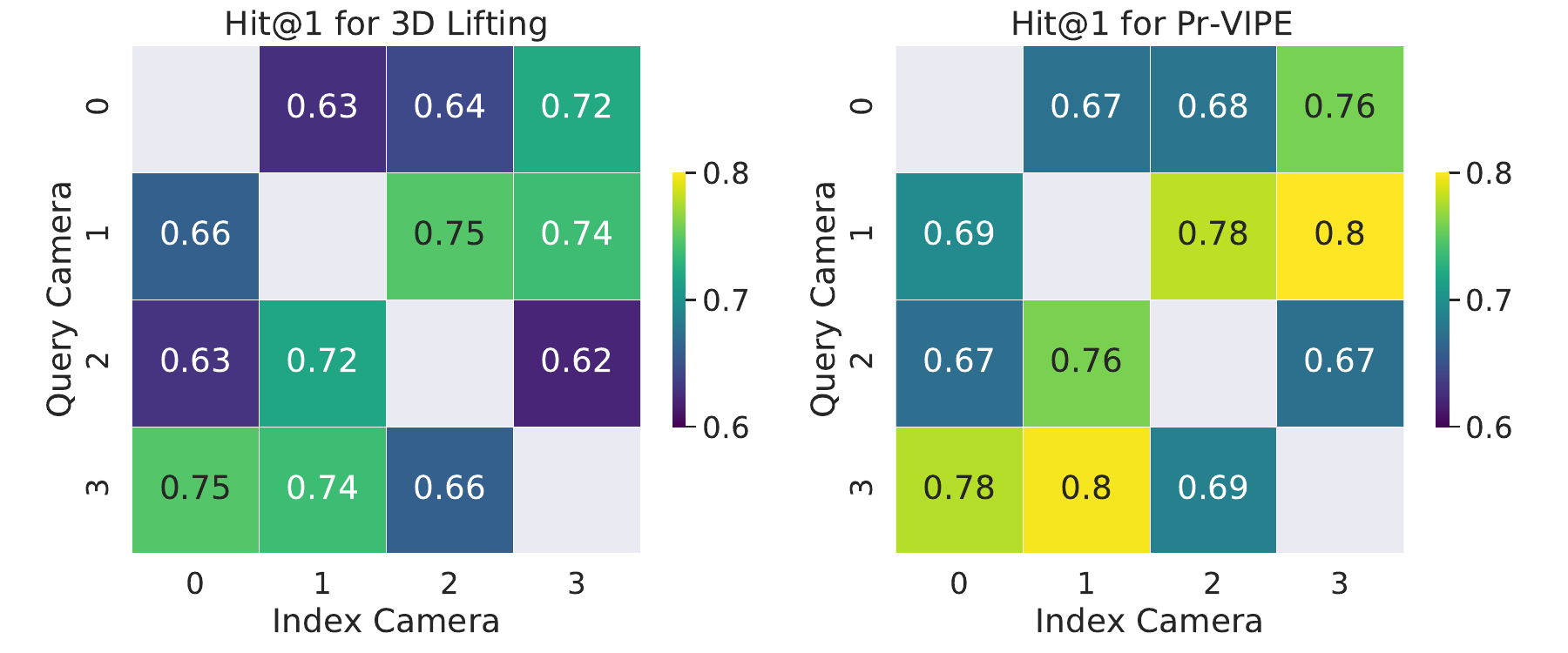}
  \caption{Comparison of retrieval Hit@$1$ of 3D lifting and Pr-VIPE (with camera augmentation) on H3.6M across different camera query and index pairs.}
 \label{fig:camera_retrieval}
\end{figure}

As described in Section~\ref{sec:retrieval_subsection}, our retrieval results are averaged over every camera query-index pair from different views. We observe that for individual query-index camera pairs, the performance is generally similar when query and index cameras are swapped, such as for H3.6M in Fig.~\ref{fig:camera_retrieval}. While there are variations in performance across different camera pairs, we observe that for all camera combinations, Pr-VIPE retrieves poses more accurately compared to the 3D lifting baseline.

\subsubsection{Effect of Sample Number $K$ and Margin Parameter $\beta$}

\begin{table}
  \centering
  \caption{Comparison of cross-view pose retrieval results Hit@$k$ ($\%$) using Pr-VIPE (with camera augmentation) with different number of samples $K$ and margin parameter $\beta$ on H3.6M.} \label{tab:ablation}
  \scalebox{0.9}{
   \begin{tabular}{c c | c c c c}
   \toprule[0.2em]
    Hyperparameter & Value & $k=1$ & $k=5$ & $k=10$ & $k=20$ \\ [0.25ex]
   \toprule[0.2em]
 \multirow{3}{*}{$K$} & $10$ & $72.6$ & $89.7$ & $93.7$ & $96.4$\\
  & $20$ & $72.9$ & $90.0$ & $93.9$ & $96.5$   \\
  & $30$ & $72.9$ & $90.1$ & $94.1$ & $96.6$ \\
 \hline
 \multirow{4}{*}{$\beta$} & $1.25$ & $72.1$ & $89.9$ & $93.8$ & $96.5$ \\
  & $1.5$ & $72.5$ & $90.0$ & $93.9$ & $96.6$ \\
  & $2$ & $72.9$ & $90.0$ & $93.9$ & $96.5$ \\
  & $3$ & $72.8$ & $89.7$ & $93.8$ & $96.4$ \\
   \bottomrule[0.1em]
\end{tabular}
}
\end{table}

Table~\ref{tab:ablation} shows the effect of the number of samples $K$ and the margin parameter $\beta$ (actual triplet ratio loss margin $\alpha=\log\beta$) on Pr-VIPE. $K$ controls how many points we sample from the embedding distribution to compute matching probability and $\beta$ controls the ratio of matching probability between matching and non-matching pairs. 

By varying $\beta$, our model performance varies slightly. Table~\ref{tab:ablation} shows that $\beta = 2$ and $\beta = 3$ has similar performance. The main effect of $\beta$ is on retrieval confidence by controlling the scaling between matching and non-matching pairs. With larger $\beta$, non-matching pairs are scaled to a smaller matching probability. 

In terms of the number of samples, Pr-VIPE performance with 10 samples is competitive, but the performance is slightly better with 20 samples. Increasing the number of samples further has similar performance. For our experiments, we use $20$ samples and $\beta = 2$.

\subsubsection{Effect of Camera Augmentation Parameters}
We explore the effect of random rotation ranges for camera augmentation on pose retrieval in Table~\ref{tab:ablation_aug}. With the same procedure as Section~\ref{sec:retrieval_subsection}, all models are trained on the 4 chest-level cameras in H3.6M, and the models with camera augmentation additionally use projected 2D keypoints from randomly rotated 3D poses. For random rotation, we always use azimuth range of $\pm180^{\circ}$, and we evaluate a range of angle limits for elevation and roll. 

\begin{table}
  \centering
  \caption{Comparison of cross-view pose retrieval results Hit@$1$ ($\%$) using Pr-VIPE with different rotation ranges for camera augmentation on H3.6M and 3DHP. The azimuth range is always $\pm180^{\circ}$ and the angle ranges in the table are for elevation and roll. The row ``w/o aug.'' corresponds to Pr-VIPE without camera augmentation.} \label{tab:ablation_aug}
  \scalebox{0.85}{
   \begin{tabular}{c c | c c c c c}
   \toprule[0.2em]
    Hyperparameter & Range & H3.6M & 3DHP (Chest) & 3DHP (All) \\ [0.25ex]
   \toprule[0.2em]
  \multirow{4}{*}{Elevation/Roll} & w/o aug. & $74.6$ & $25.7$ & $19.8$ \\
  & $\pm15^{\circ}$ & $73.4$ & $29.4$ & $25.8$  \\
  & $\pm30^{\circ}$ & $72.9$ & $29.0$ & $27.3$ \\
  & $\pm45^{\circ}$ & $72.2$ & $28.4$ & $28.0$ \\
   \bottomrule[0.1em]
\end{tabular}
}
\end{table}

The model without camera augmentation does the best on the H3.6M, which has the same 4 camera views as training. With increase in rotation angles during camera augmentation, the performance on chest-level cameras drops while performance on new camera views generally increases. Camera augmentation enables our model to generalize much better to novel views, and we use $\pm30^{\circ}$ angle range for elevation and roll.

\subsubsection{Effect of NP-MPJPE threshold $\kappa$}

\begin{table}
  \centering
  \caption{Comparison of cross-view retrieval results Hit@$1$ ($\%$) using Pr-VIPE with different NP-MPJPE threshold $\kappa$ for training and evaluation on H3.6M.} \label{tab:ablation_kappa}
  \scalebox{1.00}{
  \begin{tabular}{c | c c c c}
  \toprule[0.2em]
  & \multicolumn{4}{c}{Evaluation $\kappa$}\\
  Training $\kappa$ & $0.05$ & $0.10$ & $0.15$ & $0.20$ \\[0.25ex]
  \toprule[0.2em]
 $0.05$ & $48.3$ & $74.5$ & $89.6$ & $95.6$\\
 $0.10$ & $47.3$ & $74.6$ & $89.9$ & $95.9$ \\
 $0.15$ & $45.3$ & $73.9$ & $90.1$ & $96.1$\\
 $0.20$ & $42.8$ & $72.8$ & $89.9$ & $96.1$\\
  \bottomrule[0.1em]
\end{tabular}
}
\end{table}

We train and evaluate different NP-MPJPE thresholds $\kappa$ in Table~\ref{tab:ablation_kappa}. $\kappa$ determines the NP-MPJPE threshold for a matching pose pair, and Fig.~\ref{fig:similarity} provides a visualization of different NP-MPJPE values.

Pr-VIPE generally achieves the best accuracy for a given NP-MPJPE threshold when the model is trained with the same matching threshold. 
Table~\ref{tab:ablation_kappa} demonstrates that when we train with a tight threshold, e.g., $\kappa=0.05$, we do comparatively well at looser thresholds. In contrast, when we train with a loose threshold, e.g., $\kappa=0.2$, we do not do as well given a tighter accuracy threshold at evaluation time. This is because when we push non-matching poses using the triplet ratio loss, $\kappa=0.2$ does not explicitly push poses less than the NP-MPJPE threshold. The closest retrieved pose will then be within $0.2$ NP-MPJPE but it is not guaranteed to be within any threshold smaller than $0.2$ NP-MPJPE.

For all our other experiments, we use $\kappa=0.1$. For future applications with other matching definitions, the Pr-VIPE framework is flexible and can be trained with a different $\kappa$ or other similarity measures to satisfy different accuracy requirements.

\subsubsection{Independent Keypoint Dropout Probability}
We vary the dropout probability $q$ when training Pr-VIPE (with camera augmentation) and evaluate its effect on H3.6M retrieval with fully and partially visible poses. 
In Table~\ref{tab:ablation_random}, we see that as $q$ increases, the model performance generally improves on occluded poses but degrades on fully visible poses. For our main experiments, we choose $q=20\%$.

\begin{table}
  \centering
  \caption{Comparison of cross-view pose retrieval results Hit@$k$ ($\%$) on H3.6M with targeted occlusions using different training independent keypoint dropout probability $q$.} \label{tab:ablation_random}
  \scalebox{0.9}{
  \begin{tabular}{c c | c c c c}
  \toprule[0.2em]
    Evaluation type & $q$ ($\%$) & $k=1$ & $k=5$ & $k=10$ & $k=20$ \\ [0.25ex]
  \toprule[0.2em]
    \multirow{5}{*}{No occlusion} & $5$ & $72.5$ & $89.7$ & $93.8$ & $96.5$  \\    
    & $10$ & $72.0$ & $89.7$ & $93.7$ & $96.4$  \\    
    & $20$ & $71.7$ & $89.7$ & $93.7$ & $96.4$  \\   
    & $30$ & $71.2$ & $89.3$ & $93.5$ & $96.2$  \\   
    & $40$ & $70.1$ & $88.9$ & $93.3$ & $96.1$  \\
    \hline
    \multirow{5}{*}{Targeted occlusion} & $5$ & $61.0$ & $84.0$ & $89.8$ & $93.7$  \\    
    & $10$ & $64.5$ & $86.1$ & $91.2$ & $94.7$  \\    
    & $20$ & $66.2$ & $87.3$ & $92.1$ & $95.3$  \\   
    & $30$ & $66.7$ & $87.5$ & $92.2$ & $95.4$  \\   
    & $40$ & $66.6$ & $87.5$ & $92.3$ & $95.4$  \\    
  \bottomrule[0.1em]
\end{tabular}
}
\end{table}

%% file: 6.conclusion.tex
\section{Conclusion}

We present Pr-VIPE, a novel approach to learning view-invariant, occlusion-robust probabilistic embeddings for recognizing 3D human pose similarity using monocular 2D pose keypoints from either single images or video sequences. Our models, only trained to match similar 3D poses, achieve highly competitive performance on cross-view pose retrieval, action recognition, and video alignment tasks against baseline models trained for each task.

We further explore two synthetic occlusion augmentation strategies when training Pr-VIPE to improve its robustness to partially visible input. We show that our augmentation strategies result in significant increases in retrieval accuracy for partially visible poses. This capability of handling incomplete input enables the use of our model for realistic photos, where pose occlusions are common. It also makes it possible to devise systems for targeted partial pose search using a single embedding model.

Pr-VIPE has a simple architecture and can be potentially applied to other domains, such as hand pose or other generic object pose recognition.
With this work, we hope to encourage further explorations into approaching pose related problems from an embedding perspective, especially where recognizing 3D similarity is central to the problem.

%% file: 7.acknowledgements.tex
\section{Acknowledgements}

We would like to thank Debidatta Dwibedi, Kree Cole-McLaughlin, and Andrew Gallagher from Google Research, Xiao Zhang from University of Chicago, and Yisong Yue from Caltech for the helpful discussions. We appreciate the support of Pietro Perona and the Computational Vision Lab at Caltech for making this collaboration possible. The author Jennifer J.\ Sun is supported by NSERC (funding number PGSD3-532647-2019) and Caltech.

%% file: a1.appendix.tex
\appendix
\section{Keypoint Definition}\label{sec:appendix_keypoint_definition}
\begin{figure*}[!b]
    \centering
    \subfloat[16 3D keypoints.\label{fig:skeleton16}]{\includegraphics[width=0.4\textwidth]{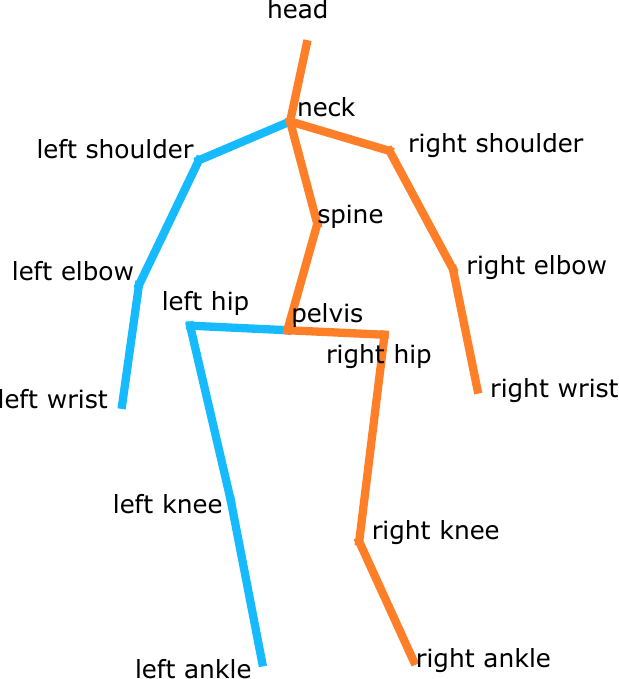} }%
    \qquad
    \subfloat[13 2D keypoints.\label{fig:skeleton13}]{\includegraphics[width=0.4\textwidth]{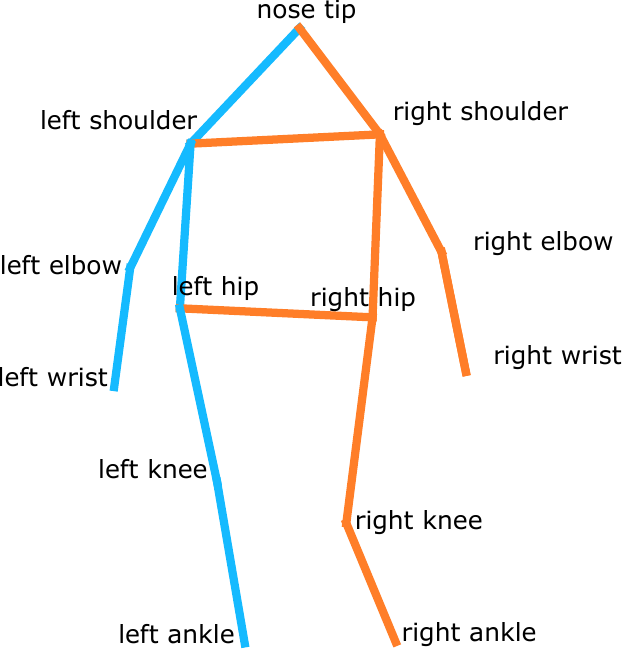} }%
    \caption{Definitions of pose keypoints.}%
    \label{fig:skeleton}%
\end{figure*}

Fig.~\ref{fig:skeleton} illustrates the keypoints that we use in our experiments.
The 16 keypoints we use to define a 3D pose are shown in Fig.~\ref{fig:skeleton16}.
We map 3D keypoints from different datasets to these 16 keypoints for training and evaluation in this paper. Besides most unambiguous mappings, several special mappings that we would like to note here are:
\begin{itemize}
    \item For the Human3.6M dataset~\cite{ionescu2013human3}, we discard the ``Neck / Nose'' keypoint and map the "Thorax" keypoint to ``Neck''.
    \item For the MPI-INF-3DHP dataset~\cite{mehta2017monocular}, we discard the ``Head top'' keypoint.
    \item For the 3DPW dataset~\cite{vonMarcard2018}, we add ``Pelvis'' keypoint as the center of ``Left hip'' and ``Right hip'', and add ``Spine'' as the center of ``Pelvis'' and ``Neck''.
\end{itemize}

The 13 2D keypoints we use to define a 2D pose are shown in Fig.~\ref{fig:skeleton13}.
We follow the COCO~\cite{lin2014microsoft} keypoint definition, keeping all the 12 body keypoints and the ``Nose'' keypoint on the head.